\documentclass[12pt,a4paper,fleqn]{cas-sc}
\usepackage{geometry}
\usepackage[numbers]{natbib}
\usepackage{amssymb}
\usepackage{amsmath}
\usepackage[ruled]{algorithm2e}
\usepackage[]{algpseudocode}
\usepackage{mathtools}
\usepackage{algcompatible}
\usepackage{algorithm2e}
\usepackage{lipsum}
\usepackage{hyperref}
\usepackage{rotating} 
\usepackage[T1]{fontenc}
\usepackage{fancyhdr}
\usepackage{supertabular,booktabs}
\usepackage{setspace}
\usepackage{tabularx}
\usepackage{longtable}
\usepackage{array,ragged2e,longtable}
\newcolumntype{P}{>{\RaggedRight\hspace{0pt}}p{\dimexpr(\textwidth-1cm -10\tabcolsep -5\arrayrulewidth)/4\relax}}
\usepackage{lscape}
\begin{document}

\begingroup 
\setlength\extrarowheight{2pt}
\setlength\LTcapwidth\textwidth
\font\myfont=cmr9 at 20pt
\title{\myfont Rule-Based Modeling of Low-Dimensional Data with PCA and Binary Particle Swarm Optimization (BPSO) in ANFIS }

\author{Afnan Al-ALi}[type=editor,
                        auid=000,bioid=1,
                       ]
 \fnmark[1]
\ead{aa1805360@qu.edu.qa}
\cortext[aa1805360@qu.edu.qa]{Corresponding author} 

\author{Uvais Qidwai}[type=editor,
                        auid=000,bioid=2,
                      ]
                      
\fnmark[1]
\credit{Conceptualization of this study, Methodology, Software}

\fnmark[1]\affiliation{organization={ Computer Science and Engineering Department},
    addressline={Qatar University}, 
    city={Doha, Qatar}
    }

\begin{abstract}
Fuzzy rule-based systems interpret data in low-dimensional domains, providing transparency and interpretability. In contrast, deep learning excels in complex tasks like image and speech recognition but is prone to overfitting in sparse, unstructured, or low-dimensional data. This interpretability is crucial in fields like healthcare and finance. Traditional rule-based systems, especially ANFIS with grid partitioning, suffer from exponential rule growth as dimensionality increases. We propose a strategic rule-reduction model that applies Principal Component Analysis (PCA) on normalized firing strengths to obtain linearly uncorrelated components. Binary Particle Swarm Optimization (BPSO) selectively refines these components, significantly reducing the number of rules while preserving precision in decision-making. A custom parameter update mechanism fine-tunes specific ANFIS layers by dynamically adjusting BPSO parameters, avoiding local minima. We validated our approach on standard UCI respiratory, keel classification/regression datasets, and a real-world ischemic stroke dataset, demonstrating adaptability and practicality. Results indicate fewer rules, shorter training, and high accuracy, underscoring the method’s effectiveness for low-dimensional interpretability and complex data scenarios. This synergy of fuzzy logic and optimization fosters robust solutions. Our method contributes a powerful framework for interpretable AI in multiple domains. It addresses dimensionality, ensuring a rule base.

\end{abstract}



\maketitle
\section{Introduction}
Fuzzy modeling is a descriptive language that a fuzzy logic system uses to represent real-world operations. These models can express expert knowledge through fuzzy if-then rules without requiring a detailed qualitative analysis. As a result, the ambiguity and uncertainty systems can be represented more transparently, enabling a thorough comprehension of the system's operation\cite{rajab2019handling}. 
Fuzzy rule-based systems are particularly adept at interpreting data in scenarios with fewer variables but complex relationships. In contrast, deep learning is renowned for its success in processing high-dimensional data, such as images, videos, and speech in different applications \cite{yan2020deep}\cite{yan2020depth}\cite{yan2021task}\cite{yan2021precise}\cite{yan2022age}\cite{yan2022review}, where it leverages vast amounts of data to uncover intricate patterns. However, deep learning techniques often face challenges when dealing with sparse, unstructured, or low-dimensional datasets due to their high parameter demands and susceptibility to overfitting. These challenges highlight the advantages of fuzzy modeling in providing clear and interpretable rules for systems characterized by uncertainty and ambiguity.
In 1965, Lotfi Zadeh introduced Fuzzy Logic as a mathematical tool to deal with uncertainty by using what is called \textit{membership functions}, such that the linguistic words like heavy, tall, low, etc., which cannot be categorized precisely in terms of “0” or “1”; instead it is possible to express them in degrees of belonging to a specific category. There are two popular types of fuzzy inference techniques: the Mamdani, which Mamdani and Assilian founded \cite{mamdani1975experiment}, and the Sugeno or Takagi–Sugeno–Kang or (TSK) fuzzy inference technique, which Sugeno pioneered \cite{takagi1985fuzzy}. The key distinction between the two approaches is the outcome of fuzzy rules. TSK fuzzy inference systems use linear functions of input variables as rule consequents, whereas Mamdani fuzzy systems use fuzzy sets as rule consequents \cite{sivanandam2007introduction}. One of the successful fuzzy inference techniques that can deal with highly complex, nonlinear systems is the Adaptive Neuro-Fuzzy Inference system (ANFIS). The typical way to write a single fuzzy rule in ANFIS based on the form of TSK inference  for two inputs $x_1$ and $x_2$ is:\\

IF $x_1$  is $A_1$  AND $x_2$ is $A_2$  THEN $y$=f($x_1$,$x_2$),  where: 
\\

$A_1$, and $A_2$ are the fuzzy sets for the input variables $x_1$ and $x_2$, y is the output, $AND$ is a logical AND gate \cite{haznedar2021optimizing}.

ANFIS  offers advantages but also faces specific challenges \cite{albertos1998fuzzy}. These challenges associated with ANFIS are interconnected, creating a complex landscape for its application. The need for expert knowledge and rule-based development is closely tied to rule-based complexity since managing and maintaining a complex rule base becomes more challenging and time-consuming. The complexity of the rule base also impacts subjectivity and interpretability, as the intricate rules may lead to inconsistencies and subjective interpretations, making it harder to ensure transparency. Furthermore, integrating ANFIS with other computational techniques presents challenges, particularly when dealing with a complex rule base and issues of subjectivity. Scalability and efficiency concerns are magnified when the rule base becomes more complex, especially in large-scale problems where resource management becomes crucial. Finally, efficient data-driven learning in ANFIS is essential to address the challenges of overfitting and model complexity, which are closely related to the complexity of the rule base. These challenges collectively highlight the intricate nature of ANFIS and the need for careful consideration and expertise in its application \cite{burda2015reduction}\cite{alonso2011hilk++}.

From the challenges delineated earlier, we can distill these complexities into three primary focal points, which we intend to concentrate on within the context of this research study, as depicted in the illustrative Figure \ref{fig2}

\begin{figure}[h]
  \centering
  \includegraphics[width=0.7\linewidth]{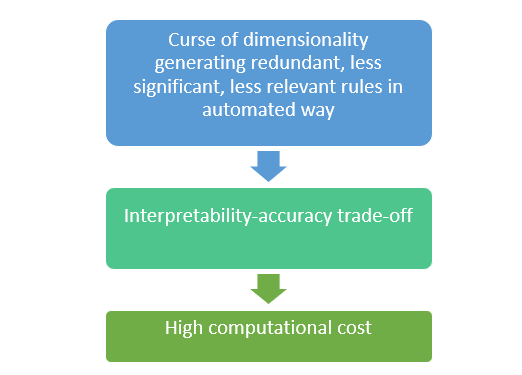}
  \caption{Main Challenges in ANFIS.}
  \label{fig2}
\end{figure}

Addressing these challenges in ANFIS requires domain expertise, careful rule-based development, thoughtful integration with other techniques, and considerations for scalability and interpretability. Additionally, advances in machine learning techniques, methodologies, and domain-specific knowledge can contribute to overcoming these challenges and enhancing the effectiveness of ANFIS in various real-world applications. The goal of this study is to develop a model that addresses the previously mentioned drawbacks and makes a balance between the tradeoff problem of interpretability and accuracy. We aim to develop a real-time and practical model that can be evaluated on real datasets effectively.

The main contributions of this work are summarized as follows:

\begin{enumerate}

    \item \textbf{Rule Reduction for Improved Efficiency:}
    Developed a novel technique to reduce the number of generated rules in ANFIS. This simplifies the model and significantly decreases training time, enhancing computational efficiency.
     \item \textbf{Two-Stage PCA-BPSO Model for rules Reduction:}
     Devised a two-stage model for advanced rule reduction:
     \begin{itemize}
         \item Stage 1: Principal Component Analysis (PCA) is applied to the firing strengths, transforming the data into principal components.
         \item Stage 2: Binary Particle Swarm Optimization (BPSO) technique selects the most informative components, ensuring minimized error.
     \end{itemize}
    This layered approach ensures comprehensive rule consideration while still achieving significant reduction.
    \item \textbf{Optimized Parameter Update Mechanism:}
    The proposed models notably reduce processing time, not just via rule reduction but also by minimizing parameter updates.
    The working loop, integrated with the PCA-BPSO mechanism, operates exclusively between layers 3 and 4 of ANFIS. Consequently, only the consequent parameters are updated during the forward path in each iteration.
    \item \textbf{Dynamic BPSO Parameters Update }
    The BPSO parameters represented by the inertia weight and the acceleration coefficients are dynamically adjusted based on swarm performance to prevent stagnation in local minima and ensure a balanced exploration-exploitation trade-off. 
    \item \textbf{Versatile Application:}
    Demonstrated the versatility of the proposed models by applying them to diverse tasks, including classification and regression.
    As a testament to its real-world applicability, the model was successfully employed on a real dataset of ischemic stroke.
    
\end{enumerate}
 
The rest of the paper is organized as follows: section 2 explains the ANFIS architecture in detail, and section 3  presents a comprehensive review of ANFIS optimization to solve the trade-off problem of interpretability- and accuracy. Section 4 presents the methodology, Section 5 shows the experimental setup, and the results and our other ablations are detailed in Sections 6 and 7, respectively. Section 8 discusses the results we obtained. Section 9 is the application of our model on the real dataset. Finally, Section 10 is the conclusion. 

\section{Adaptive Neuro-Fuzzy Inference System (ANFIS)}
ANFIS is a powerful estimating model that may be used with various machine learning methods in addition to neuro-fuzzy systems. Despite being well-liked by academics, ANFIS has drawbacks, such as the curse of dimensionality and computing expense, that prevent it from being applied to situations with massive inputs. There is a lot of potential for improvement, even though several techniques have been put forth in the literature to address these flaws \cite{salleh2017adaptive}.
ANFIS structure mainly consists of two main parts: the antecedent or input part and the consequent or the results part. These two sections form the fuzzy rules and the network's final shape. The parameters of each section are updated during the training process based on hybrid optimization technique \cite{haznedar2021optimizing}. The entire network of ANFIS consists of five layers, as shown in Figure \ref{fig_2}. The antecedent part of its network includes layers $1 - 3$, while the consequent part includes the rest.
The first layer is the fuzzification layer, where the membership degrees for each input are calculated in the formulas below where \textbf{L} refer to the layer associated with its order, and \textit{O} is the output of the layer, where, the first number refer to the layer order and the second number refer to the input order:\\
$\mathbf{L1-L2}$:
\begin{equation}
\hspace{3cm} 
 O_{1_1}=\mu_{A_i}(x_1);  O_{1_2}=\mu_{B_i}(x_2);   
\end{equation}
\begin{figure*}[!t]
\centering
\includegraphics[width=0.85\textwidth]{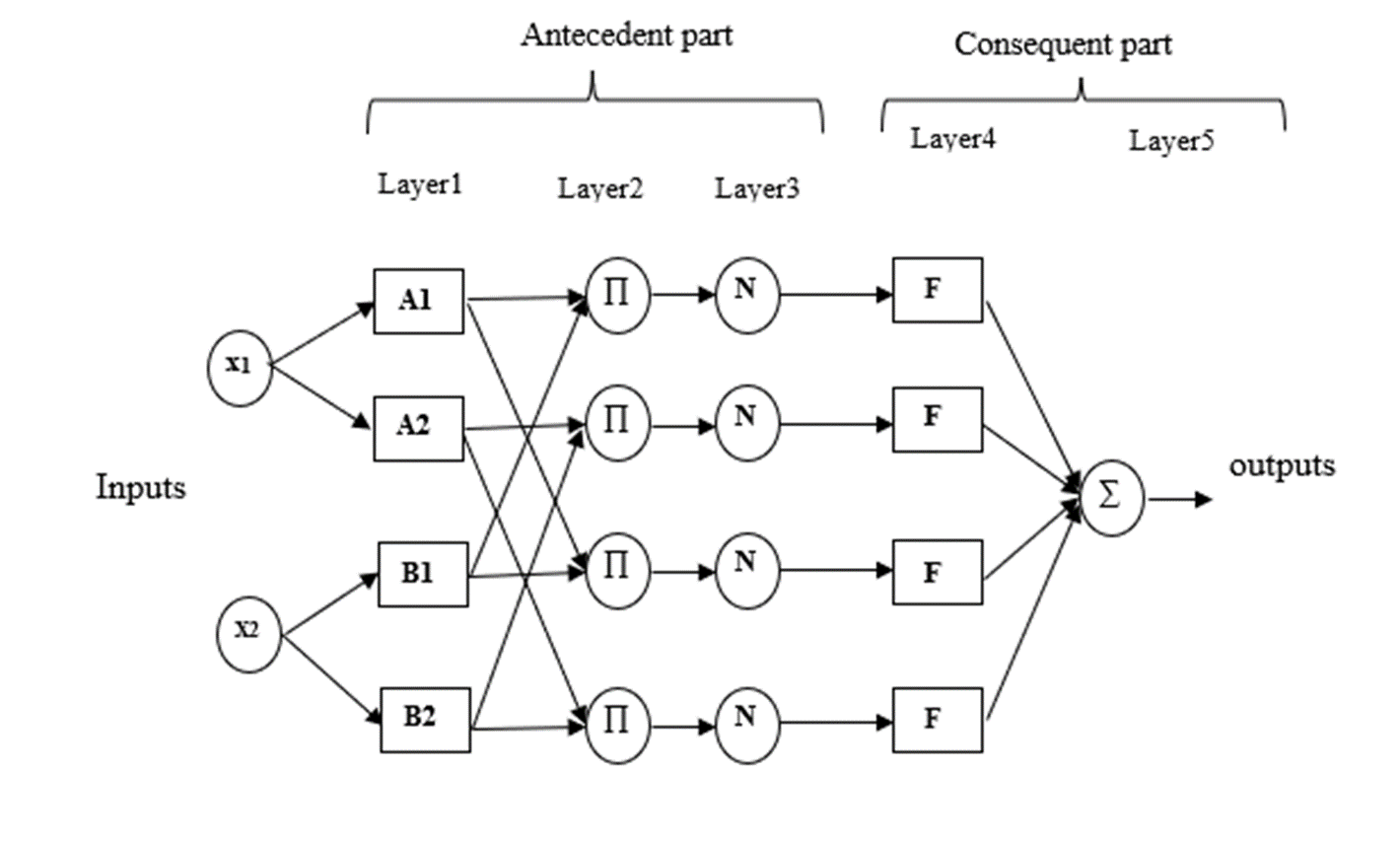}
\caption{Detailed Traditional ANFIS Architecture with Five Distinct Layers, where the squared nodes refer to the adaptive nodes and the circle nodes refer to the fixed nodes.}
\label{fig_2} 
\end{figure*}
where $\mu_{A_i}(x_1)$ and $\mu_{B_i}(x_2)$ are the membership functions of each input. The nodes of this layer are adaptive (in square shape), which means that its parameters are updated in a backward path using the gradient descent algorithm. The membership functions can take different forms depending on the dataset and the process. The Gaussian and Generalized bell-shaped membership functions are the most popular types \cite{dombi1990membership} \cite{talpur2017investigation}.
The second and third layers are the rule and normalization, respectively. Both have fixed nodes (circle shape). The ruling layer calculates the firing strength of each rule by multiplying the membership values of the previous layer for a combination of inputs as equation 2 or by taking the minimum value of them (the t-norm). The third layer normalizes these firing strengths \textcolor{red}{$w$} using the min-max normalization technique as shown in equation 3. \\
$\mathbf{L2-L3}$:
\begin{equation}
\hspace{3cm} 
 O_2,j=w_j=\mu_{A_i}(x_1)* \mu_{B_i}(x_2)\\ 
\end{equation}
$\mathbf{L3-L4}$:
\begin{equation}
\hspace{3cm} 
  O_3,j=\overline{w_j}=\frac{w_j}{\sum_j{w_j}}\\ 
\end{equation}
The fourth layer is the beginning of the consequent part, known as the defuzzification layer, see equation 4, representing the weighted values of the rules using a first-order polynomial equation of consequent parameters multiplied by the normalized firing strengths. This layer consists of adaptive nodes whose parameters (the consequent parameters) are updated in the forward path using the least square error technique.
$\mathbf{L4-L5}$:
\begin{equation}
\hspace{3cm} 
O_4,j=\overline{w_j}f_j=\overline{w_j}(p_jx_1+q_jx_2+r_1)\\ 
\end{equation}
Here $p_j, q_j, r_1$ are the consequent parameters of each rule $j$, which they equal 
\begin{equation}
\hspace{3cm} 
\begin{bmatrix}
p \\
q \\
r \\
\end{bmatrix}
=
\left( n_{\text{inputs}} + 1 \right) \times j
\end{equation}
where $n_{inputs}$ is the number of inputs \cite{karaboga2019adaptive}.
These two types of updates of ANFIS parameters are known as the two-pass hybrid learning algorithm. Finally, the fifth layer is the output layer, which is the summation of all the outputs of layer four. 
$\mathbf{L5}$:
\begin{equation}
\hspace{4cm} 
  O_5=\sum_j{\overline{w_j}f_j}\\ 
\end{equation}
The nodes of each layer are connected to the next layer by directed links. Therefore, to produce the output for a single node, each node performs a particular function on its incoming signals \cite{talpur2020deep}.

\section{Related work}\label{related2}
As we mentioned earlier, there is a trade-off problem with ANFIS related to its interpretability and accuracy. The ability of fuzzy models to describe their systems' patterns is called interpretability. Some studies concurred that the model structure, the number of input variables, the number of fuzzy rules, the number of linguistic words, and the form of the fuzzy sets are all aspects of interpretability. Because they affect the system's complexity and time consumption. The definition of accuracy relates to the fuzzy model's ability to describe the system being modeled accurately \cite{rini2013balanced}. Numerous attempts have addressed ANFIS issues by enhancing either its interpretability or accuracy, or both using optimization techniques. We have explored the literature and discovered that researchers have pursued two main avenues to resolve these problems, which we will elaborate on in the subsequent sub-sections.
\subsection{Issues with ANFIS Training and Overfitting}
The training of ANFIS architecture constitutes an optimization process to find the best values for its antecedent and consequent parameters. Commonly used derivative-based learning algorithms for this purpose include the Levenberg Marquardt (LM) \cite{ranganathan2004levenberg}, backpropagation (BP)\cite{rumelhart1995backpropagation}, Kalman filter (KF) \cite{welch1995introduction}, and gradient descent (GD) algorithms \cite{baldi1995gradient}. These, however, carry the potential risk of local minimum problems due to the chain rule, and calculating the gradient at each step can be challenging. Additionally, the effectiveness of these algorithms heavily depends on the initial values, and the convergence of the parameters can be relatively slow. Recently, several ANFIS studies have substituted these learning algorithms with evolution optimization or metaheuristic optimization algorithms, including the Genetic Algorithm (GA) \cite{whitley1994genetic}, Differential Evolution (DE) \cite{price2013differential}, Artificial Bee Colony (ABC) \cite{karaboga2010artificial}, and Particle Swarm Optimization (PSO) \cite{chen2013hybrid}.

In \cite{kurniawan2018premise}, the researchers proposed a combination of particle swarm optimization and genetic algorithm. Although PSO is known for its robustness and fast solving of non-linear problems, its quickness could lead to local optimum solution space convergence. To tackle this problem, the researchers merged the GA algorithm with PSO to develop ANFIS-PSOGA for optimal premise parameters.

In \cite{ebrahimi2021accuracy}, a more specific and interpretable fuzzy model (ANFIS-BAT) was proposed for predicting dust concentration in both cold and warm months across semi-arid regions of Iran. The researchers employed the bat algorithm for fine-tuning the premise and consequent parameters of the ANFIS network to minimize the cost function in the learning process. Another training algorithm, BWOA-ANFIS \cite{tightiz2020intelligent}, was proposed to replace the gradient descent in traditional ANFIS. The authors of this study applied the association rules learning technique (ARLT) and then tuned the premise and consequent parameters by utilizing the Black Widow Optimization Algorithm (BWOA).
In the study of  \cite{rajeshwari2022dermatology}, the authors introduced an ANFIS-FA methodology. This system utilized ANFIS  combined with subtractive clustering (SC). This model's unique aspect was using a firefly optimization technique (FA) to improve the optimization of all SC parameters. These parameters, which included the range of influence, squash ratio, accept ratio, and reject ratio, were explicitly optimized to enhance the system's ability to classify and diagnose skin cancer at its early stages.

In \cite{salleh2017optimization}, both the premise and consequent parameters were optimized using the artificial bee colony (ABC) optimization algorithm to enhance the precision of ANFIS in classifying Malaysian SMEs. The ABC algorithm was employed to update these parameters in forward and backward passes instead of the traditional hybrid learning algorithms in conventional ANFIS. Although this technique demonstrated high accuracy, the ABC requires a more efficient exploration strategy.

\subsection{Issues with ANFIS Interpretability and Complexity}\label{related2}

To enhance the interpretability of ANFIS, researchers have attempted to optimize the rule base using various techniques, such as reducing the number of features using feature selection techniques or eliminating redundant rules using different pruning methods.

The process of reducing the number of input variables leads to a lesser number of system parameters and, thus, a reduced number of generated rules. Consequently, this results in a more interpretable model and a less complex ANFIS structure. Several techniques are available for the feature selection, including traditional techniques or filter methods that depend on distance measurement and redundancy features \cite{li2022feature}\cite{saberi2022dual}. Other strategies known as wrapper methods utilize evolutionary algorithms to evaluate the best-selected features, such as binary particle swarm optimization (BPSO), ant colony optimization (ACO), and GA. A classifier is used as an objective function to calculate the minimum error of each subset.
In \cite{rahchamani2021hybrid}, a feature selection based on a genetic algorithm was proposed as the main contribution to simplifying ANFIS complexity to reach high performance by reducing the feature vector in the concrete production industry.
In the study of \cite{birgani2019optimization}, the accuracy of ANFIS was significantly improved while maintaining a less complicated architecture through feature selection. They utilized Principal Component Analysis (PCA) to classify brain tumor MRI images. The efficiency of their technique was demonstrated through the generation of fewer fuzzy rules and an improvement in system accuracy. This was achieved by integrating image segmentation with thresholding techniques and increasing the number of iterations.
Another similar approach is dividing the input variables based on information granules such as fuzzy sets, then using the Apriori algorithm to create an initial rule base that efficiently derives high-quality, interpretable rules from vast datasets. These rules were concurrently selected and tuned to improve the model's performance, as seen in \cite{fazzolari2014multi}, and a similar idea is in \cite{antonelli2016multi}.

Optimizing the rule base based on rule growing and pruning techniques has been implemented to minimize the rule base while maximizing accuracy. The rule base is a crucial part of any fuzzy inference system (FIS), and the quality of its results hinges on the efficacy of these rules. Not all generated rules are essential or contribute significantly to improving accuracy. Many of them are inefficient and can be pruned to reduce the complexity of the FIS system \cite{rini2013balanced}\cite{hussain2015analysis}.

Several techniques have been proposed for rule-based optimization. One such technique is \textit{clustering}, as in \cite{leonori2020generalized}, where various clustering techniques ((one partitional and two other hierarchical strategies) have been proposed to synthesize ANFIS. They addressed the issue of membership function overlap by considering the input space. After the clustering process, a Min-Max classifier is used to refine the membership function definitions. This process generates a limited number of rules while ensuring coverage of the entire input space.
In \cite{suraj2016jaya}, a different clustering technique was adopted to expedite training time, prune irrelevant rules, and enhance the accuracy of classifying motor imagery tasks for controlling light-emitting diodes. Their methodology involved splitting the dataset into two clusters using the k-means clustering algorithm and triggering rules based on each cluster. The Jaya algorithm was combined with ANFIS to determine each group's optimal number of features, subsequently informing the rules' triggering.

Another clustering technique was used in \cite{pramod2021k} to improve the interpretability of grid partitioning ANFIS. This technique proposed K-Means clustering-based Extreme Learning ANFIS (KMELANFIS) for regression purposes. The input space was clustered, and the clustering centers were used to initialize the membership function parameters. The membership functions were reduced using a similarity index method between adjacent ones. Finally, an extreme learning machine (ELM) was used to compute the consequent parameters. However, this technique is limited when applied to a low-dimensional dataset. 

Another approach to rules pruning is to use \textit{thresholding techniques}, as seen in \cite{rini2013balanced} and  \cite{wang2012assessment}, where the elimination of non-essential rules was achieved by employing a threshold set by an expert. All rules with firing strengths below this threshold were discarded. However, these methodologies, reliant on human experts for threshold determination, faced challenges. Especially in cases where expert opinions conflicted based on data types or specific application requirements, choosing the correct threshold value proved problematic. Moreover, choosing to prune rules based on a pre-selected threshold could risk removing some significant rules, negatively impacting the system's accuracy. One possible solution is to use an adaptive threshold, as in \cite{guendouzi2021new}, where the authors used a new threshold-based fitness function that is adaptive.
As outlined in \cite{owoseni2020improved}, an enhanced ANFIS technique leverages a probability trajectory and a k-nearest neighbor-based clustering ensemble to refine its rules. This approach empirically establishes a threshold to select the optimal number of clusters, which can confine this method to a specific dataset type.

A solution to the rules explosion problem, Patch Learning (PL), was presented in \cite{huang2022fuzzy}. Despite its effectiveness, this technique may increase the number of patches, leading to heightened system complexity.

In \cite{held2006extracting}, a unique solution was proposed. The authors designed a three-layer ANFIS architecture for healthy infant sleep classification, utilizing a simple rule-elimination process to balance interpretability and accuracy. They modified the third layer of ANFIS to correspond with the five classes in the fifth layer of conventional ANFIS. Each node in the third layer performed a weighted sum operation of the incoming rules' firing strengths, later altered by the sigmoid function. Consequently, any rule with a normalized average contribution lower than an empirically chosen 7\% threshold was pruned due to its insignificant contribution to the classification.
Moreover, they further streamlined the rules by merging those sharing the same output class and only differing in the fuzzy concept linked with one pattern. A similar rule combination and feature selection methodology were employed in \cite{feng2020accuracy}, where the researchers employed the CFBLS model, which uses a single TSK fuzzy system, streamlining rule interpretation. The input space in CFBLS is uniformly partitioned, typically into 2, 4, or 6 parts, for better data coverage. They adopted a random selection method for features and rules to counter the "rule explosion" arising from numerous features, using a rule-combination matrix and a "don't care" matrix. Parameters were optimized through a ten-fold cross-validation coupled with a grid search. After conducting experiments 30 times per dataset, their approach aimed to harmonize accuracy and complexity in fuzzy learning systems.

In \cite{do2021prediction}, rule optimization was achieved by minimizing the node count, which correlates with the number of generated rules. The authors used a "rule-drop" technique that randomly activated and deactivated nodes in the fuzzification layer during each training step. The choice of nodes was based on probability values that served as a hyperparameter to retain a neuron within the network.
Some techniques, like the one in \cite{huhn2009furia}, bypass the pruning phase and directly learn the first set of rules from the entire training set. This direct learning from data allows for the rule's antecedents to be learned, and if the rule has no antecedent, a default rule is generated. As a result, there is a set of rules for each class.

Some other techniques restrict the number of generated rules, which is close to the thresholding approach, such as in  \cite{tomasiello2022fractional} where the authors proposed a rule reduction technique for regression purposes by using the least-squares method with fractional Tikhonov regularization, and the number of rules restricted to the number of fuzzy terms for each variable resulting in a simplified rule base that efficiently manages high-dimensional inputs while maintaining accuracy.
Another type of thresholding attempts in \cite{alcala2011fuzzy}, where the authors used a search tree for rules generation to list all possible fuzzy item sets of a class, with attributes having an order, and utilizing support thresholds to limit rule expansions; then the candidate Rule Prescreening for subgroup discovery has been conducted to select the most interesting rules by weighting patterns to ensure diverse rule coverage; and finally the genetic post-processing for rule selection and parameter tuning. 

Apart from these techniques, there are several other proposed solutions to improve the interpretability of ANFIS, such as similarity analysis \cite{rajab2019handling}, frequent pattern mining \cite{marimuthu2019oafpm}, and equalization of fuzzy rules with the membership functions\cite{tomasiello2022fractional} that also managed to achieve a degree of an interpretable framework.

In the existing literature, several notable voids have been identified in the context of optimizing ANFIS models:
\begin{itemize}
    \item Feature Selection and Rule Generation Trade-off: Prior research has focused on feature selection techniques to improve ANFIS performance. However, a critical concern is the trade-off between reducing features and maintaining effective rule generation. When reducing features, the number of rules generated might decrease, potentially leading to the exclusion of crucial rules and impacting accuracy. This study takes a distinct approach by using a complete set of features to generate a comprehensive rule set, aiming to balance interpretability and accuracy.

    \item Limitations of Clustering Techniques in ANFIS Pruning: While clustering techniques have shown promise in ANFIS rule pruning, there are still open questions regarding their limitations, especially hierarchical clustering. Issues like sensitivity to data point ordering, scalability concerns, and imbalanced cluster generation need careful consideration. This research introduces grid partitioning as an alternative to hierarchical clustering by uniformly partitioning the relevant domain to address these drawbacks and enhance the ANFIS rule pruning quality.

     \item Subjectivity and Challenges in Thresholding Techniques: Existing literature often relies on subjective expert opinions to select threshold values for ANFIS rule pruning, raising concerns about applicability across diverse data types and domains. This gap highlights the need for objective thresholding techniques that determine optimal pruning thresholds based on data characteristics. The study introduces Binary Particle Swarm Optimization (BPSO) to address this subjectivity concern, enhancing the precision and adaptability of thresholding techniques in ANFIS optimization.

    \item Firing Strengths as an Automatic Rule Pruning Metric: The literature indicates that utilizing firing strengths for rule pruning in ANFIS has been relatively underexplored. Firing strengths within ANFIS offer insights into the significance of individual rules. There is an opportunity to develop techniques that leverage firing strengths as an automatic rule-pruning metric, potentially improving the interpretability and accuracy of ANFIS models.
\end{itemize}

\section{Methodology}
This section will explain our proposed model by explaining each part embedded in the ANFIS architecture and its function.
\subsection{Principal Component Analysis (PCA)}
PCA is a prominent data science and machine learning technique for dimensionality reduction \cite{hasan2021review}. It aims to simplify complex datasets by transforming them into a lower-dimensional space while preserving essential information \cite{daffertshofer2004pca}. PCA identifies principal components, linear combinations of original features designed to capture the highest data variance, and is orthogonal to ensure variables' uncorrelation. It streamlines datasets by discarding less significant components, offering advantages such as more manageable data visualization, alleviating challenges associated with high-dimensional data, and mitigating the curse of dimensionality \cite{demvsar2013principal}. Moreover, PCA is an effective noise-reduction tool by eliminating irrelevant features, resulting in a more focused and informative dataset, ultimately enhancing the performance of machine learning algorithms \cite{ kirisci2019anfis}

Let's denote the dataset as \(X\) with \(N\) data points,  \(D\) features, and \( y \) be the labels. The goal is to reduce the dimensionality from \(D\) to \(K\) (\(K < D\)) using PCA \cite{demvsar2013principal}. Then, after ignoring the labels: ${D} = [X, y]$, ${X_{new}} = {X}$\\
Then the following steps will be followed \cite{jolliffe2016principal}:

First, Calculate the average for each dimension across the entire dataset:
    Assuming \( \mathbf{A} \) is a matrix has \( n \) rows (data points) and \( d \) columns (features or dimensions), the mean vector \( \boldsymbol{\mu} \) is:
\begin{equation}
 \hspace{3cm}
\mu_j = \frac{1}{n} \sum_{i=1}^{n} a_{ij} \quad \text{for } j = 1, 2, \ldots, d
\end{equation}
where \( a_{ij} \) is the element in the \( i \)-th row and \( j \)-th column of \( \mathbf{A} \), and \( \mu_j \) is the mean of the \( j \)-th column (or \( j \)-th feature).
The mean vector \( \boldsymbol{\mu} \) is then:
$\boldsymbol{\mu} = \left[ \mu_1, \mu_2, \ldots, \mu_d \right]$

Then, the covariance between two variables \(X\) and \(Y\) is given by:
    \begin{equation}
     \hspace{3cm}
    Cov (X, Y) = \frac{\sum_{i=1}^{n} (X_i - \mu_X)(Y_i - \mu_Y)}{n} \end{equation}
    For the dataset matrix \( \mathbf{A} \) with \( d \) features, the covariance matrix \( \Sigma \) is of size \( d \times d \). Each entry \(\Sigma_{ij}\) in this matrix represents the covariance between the \(i\)-th and \(j\)-th features:
   \begin{equation}
    \hspace{4cm}
   \Sigma = \text{Cov}(\mathbf{A})
   \end{equation}
    
Later, the eigenvectors and their corresponding eigenvalues have to be determined:
    Given matrix \( \mathbf{A} \), the eigenvalues \( \lambda \) and eigenvectors \( \mathbf{v} \) satisfy:
     \begin{equation}
      \hspace{4cm}
      \mathbf{A}\mathbf{v} = \lambda \mathbf{v} \end{equation}
    The eigenvalues are the solutions to the characteristic equation:
    \begin{equation}
     \hspace{4cm}
    \text{det}(\mathbf{A} - \lambda \mathbf{I}) = 0  \end{equation}
    $I$ is an identity matrix.
Then select \(k\) eigenvectors with the highest eigenvalues to form a \(d \times k\) dimensional matrix \( \mathbf{W} \):
    Let \( \lambda_1, \lambda_2, ... \) be the sorted eigenvalues in decreasing order. The corresponding eigenvectors are \( \mathbf{v}_1, \mathbf{v}_2, ... \). The matrix \( \mathbf{W} \) is formed by taking the first \(k\) eigenvectors: $\mathbf{W} = [\mathbf{v}_1, \mathbf{v}_2, ..., \mathbf{v}_k]$.
    
After that, Mapping the samples to the newly defined subspace:
    Given the data matrix \( \mathbf{X} \) and the eigenvector matrix \( \mathbf{W} \), the transformed data \( \mathbf{Y} \) in the new subspace is given by:
  \begin{equation}
  \hspace{4cm}
  \mathbf{Y} = \mathbf{X} \times \mathbf{W}^T  \end{equation}
\subsection{Binary Particle Swarm Optimization Technique (BPSO)}

The BPSO Technique is a discrete version of the original Particle swarm optimization algorithm proposed by Kennedy and Eberhart in 1997 \cite{kennedy1997discrete}. Its concept is the same as inspired by the physical movement of a fish school or bird flock when trying to get food, find partners, or avoid attackers \cite{dhindsa2022binary}. Like PSO, in BPSO, each particle represents a possible solution, and a population is a group of particles. Each particle has two parameters, the velocity $v$ and the position $x$, and each parameter involves the personal best $(P_{best})$ and global best $(G_{best})$ solutions in their update \cite{informatics6020021}.

For each particle in BPSO: The velocity $v$ is updated as the following equation:
\begin{equation}
\begin{aligned}
v_{i,d}(t+1) &= w.v_{i,d}(t)+ c_1 rand_1(P_{best}-x_{i,d}(t)) +\\ &\quad c_2 rand_2(G_{best}-x_{i,d}(t))
\end{aligned}
\end{equation}

$i$ is an identifier for individual particles within a swarm, with $N$ particles in total. Each particle has $D$ dimensions, represented by the index $d$, which allows us to specify each dimension uniquely, $v_{i,d}$ the velocity of particle $i$, $x_{i,d}$ is the particle's position, $w$ refers to the inertia weight, and $c_1$ and $c_2$ denote the personal and social learning factors, respectively. Random values between 0 and 1 (from a uniform distribution) are represented by $rand_1$ and $rand_2$, and $t$ is the number of iterations. The swarm's best global and personal positions are denoted by $G_{best}$ and $P_{best}$, respectively.

Then, this velocity is converted into the probability value using a Sigmoid function as shown below:
\begin{equation}
 Sig(v_{i,d}(t+1))=\frac{1}{1+\exp{-v_{i,d}(t+1)}}
\end{equation}
The position will be updated as the equation below:
\begin{equation}
x_{i,d}(t+1)=
    \begin{cases}
     1, & \text{if}\ rand_3<Sig(v_{i,d}(t+1)) \\
     0, & \text{otherwise}
    \end{cases}
\end{equation}
 $ rand_3 ()$ is a random number of uniform distribution within the range [0,1].
As each iteration unfolds, $P_{best}$ and $G_{best}$ are updated to transfer the particle exhibiting the minimum fitness function into the following iterations. $P_{best}$ and $G_{best}$ are updated according to the equations \eqref{theeq8}, and \eqref{theeq9} respectively:
\begin{equation}
\resizebox{0.7\textwidth}{!}{$P_{best,i}(t+1)=
    \begin{cases}
     x_i(t+1), & \text{if}\ F(x_i(t+1)) < F(P_{best,i}(t)) \\
     P_{best,i}(t), & \text{otherwise}
    \end{cases}$}
    \label{theeq8}
\end{equation}
\begin{equation}
\resizebox{0.7\textwidth}{!}{$G_{best}(t+1)=
    \begin{cases}
     P_{best,i}(t+1), & \text{if}\ F(P_{best,i}(t+1))< F(G_{best}(t)) \\
     G_{best}(t), & \text{otherwise}
    \end{cases}$}
    \label{theeq9}
\end{equation}
where $F(.)$ is the objective function \cite{informatics6020021} \cite{deif2022diagnosis}.
Even though numerous feature selection algorithms have been put forth, most are plagued by either high computing costs brought on by a wide search space or issues with a standstill in the local optima. Therefore, to handle feature selection problems, an effective global search approach is required \cite{dhindsa2022binary}. Meta-heuristic algorithms are considered successful candidates to achieve this goal. Among the available techniques of this type, BPSO is one of the most extensively employed due to its simple implementation, fast convergence, and low computation cost \cite{informatics6020021}. In \cite{too2019emg}, pbest-guide binary particle
swarm optimization (PBPSO) was proposed for selecting the optimal set of EMG features to improve classification performance where it reduces up to 90\% of the features keeping the high accuracy. Another study in \cite{dhindsa2022binary} where BPSO for features selection was examined to predict the class of knee angle. In their study, the BPSO reduced the features to 30\% of the total group to achieve an accuracy of 90\%. BPSO was also used to select the best set among features extracted by several deep learning models applied on histopathological images in \cite{deif2022diagnosis} to predict oral cavity squamous cell carcinoma at a low cost. In \cite{too2019new}, a new co-evolution binary particle swarm optimization technique was applied for classification purposes on ten UCI learning respiratory dataset benchmarks as a feature selection technique. They compared their technique with other feature selection methods and approved its effectiveness and ability to be applied for various applications in engineering.

\subsection{ANFIS-PCA-BPSO based rules Reduction}
Inspired by the influential role of PCA and its incorporation with ANFIS for feature reduction, as demonstrated in previous studies such as in \cite{kirisci2019anfis}\cite{sharma2015new}\cite{caesarendra2017emg}, and the effectiveness of using BPSO as a features selection technique, as mentioned in the previous subsection.
We attempt to integrate the PCA in this scenario of our methodology by allowing the firing strengths to undergo additional reduction. These firing strengths within ANFIS indicate the impact of inputs on the outputs, encapsulating the significance of the rules. Consequently, these firing strengths are crucial in compensating for internal features. Subsequently, the obtained components were optimized using the Binary Particle Swarm Optimization Technique. This particular optimization technique was utilized for selecting the optimal components of the rules, aiming to minimize the error based on the designated objective function in the context of classification or regression tasks. Figure \ref{fig8} shows the block diagram of the proposed model.
\begin{figure*}[h]
  \centering
  \includegraphics[width=0.95\linewidth]{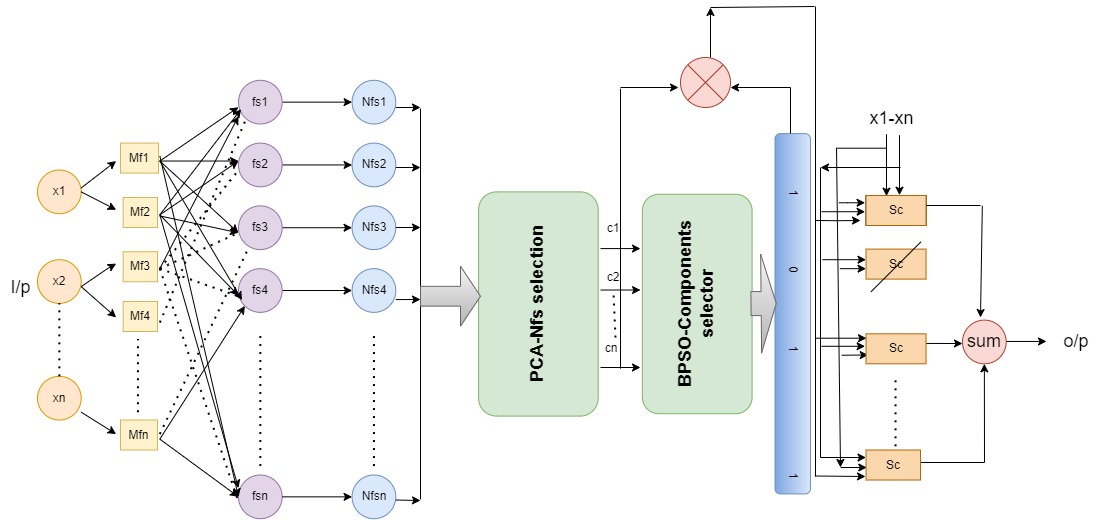}
  \caption{The architecture of the proposed ANFIS-PCA-BPSO, showing the addition of the two stages of PCA and BPSO between layers 3 and 4.}
  \label{fig8}
\end{figure*}
To explain how each part of our methodology contributes to the results, the following subsections are explaining this in detail.
\subsubsection{ANFIS-PCA for Reducing Normalized Firing Strengths}
Considering the ANFIS training process, then mathematically, let's denote the matrix of normalized firing strengths as:
\begin{equation*}
\hspace{4cm}
   MAT_{N_{fs}} \in \mathbb{R}^{N \times M} 
\end{equation*}

where \( N \) is the number of samples and \( M \) is the number of rules.

$MAT_{N_{fs}}$ = $\begin{bmatrix} Nfs_1^1 & Nfs_1^2 & ...&Nfs_1^{m-1}&Nfs_1^m\\ Nfs_2^1 & Nfs_2^2 & ...&Nfs_2^{m-1}&Nfs_2^m\\ : & : & ...&:&:\\ : & : & ...&:&:\\ : & : & ...&:&:\\ Nfs_n^1 & Nfs_n^2 & ...&Nfs_n^{m-1} & Nfs_n^m \end{bmatrix}$\\

BPSO will generate another matrix considered
a switch to select the rules. It can be represented
in the discrete form of ones and zeros; let's define this matrix as (keys) 
\begin{equation*}
\hspace{4cm}
 \text{keys} \in \{0,1\}^{M \times D}   
\end{equation*}

Where \( D \) is the number of features and \( M \) is equal to the number of rules.
$keys$ matrix is shown below:\\

$keys$ = $\begin{bmatrix} 1_1^1 & 0_1^2 & ...&0_1^{d-1}&1_1^d\\ 0_2^1 &1_2^2 & ...&1_2^{d-1}&1_2^d\\ : & : & ...&:&:\\ : & : & ...&:&:\\ : & : & ...&:&:\\ 1_m^1 & 0_m^2 & ...&1_m^{d-1} & 0_m^d\\ \end{bmatrix}$\\
Let 
\begin{equation*}
\hspace{3cm}
 f: \mathbb{R}^{N \times M} \rightarrow \{0,1\}^{M \times D}   
\end{equation*}

 be a function that maps the normalized firing strengths to the BPSO-generated keys. Then, the integration can be defined as:
\begin{equation}
\hspace{3cm}
 MAT_{\text{result}} = f(MAT_{N_{fs}})   
\end{equation}

The resultant matrix from the product of the two matrices can be denoted as:
\begin{equation}
\hspace{3cm}
MAT_{\text{result}} = MAT_{N_{fs}} \odot \text{keys}
\end{equation}
Where \( \odot \) represents element-wise multiplication (Hadamard product).

This product represents a set of candidate rules to proceed to the following layers for evaluation of size $M x D$\\

$\prod MAT_{N_{fs}} \ keys$ = 
$\begin{bmatrix} Nfs_1^1 & 0_1^2 & ...&0_1^{d-1}&Nfs_1^d\\ 0_2^1 & Nfs_2^2 & ...&Nfs_2^{d-1}&Nfs_2^d\\ : & : & ...&:&:\\ : & : & ...&:&:\\ : & : & ...&:&:\\ Nfs_m^1 & 0_m^2& ...&Nfs_m^{d-1} & 0_m^d\\ \end{bmatrix}$\\

Drawing on our understanding and insights gained from the literature, it has been observed that both Binary Particle Swarm Optimization (BPSO) and its continuous version, Particle Swarm Optimization (PSO), predominantly employ the K-nearest neighbors (KNN) algorithm as a classifier during the feature selection process.
Our proposed model has integrated BPSO as an embedded technique within the ANFIS classifier structure. In this approach, the selected components of the normalized firing strengths are treated as input features to the BPSO algorithm, generating a matrix comprising candidate rules. This integration enables the joint optimization of feature selection and rule generation within the ANFIS framework. Selecting the number of PCA components is based on the rule variance. It is usually \% 95 of the total components to retrieve the most significant amount of the data, and only a few will be removed. Algorithm 1 shows how to call the PCA-BPSO during the ANFIS training. 
\subsubsection{BPSO Parameters' update for Preserving Model Performance}
Regarding the BPSO parameters, the inertia weight, and the acceleration parameters, 
the adaptive approach of updating their values provides better adaptability, potentially improving performance by dynamically balancing exploration and exploitation.  This flexibility can be particularly
beneficial across varied problem landscapes \cite{bansal2011inertia} \cite{tian2019chaotic}.

Regarding the inertia weight, There are many inertia updating techniques such as Random \cite{eberhart2001tracking}, adaptive \cite{kessentini2015particle}, linear \cite{xin2009particle}, sigmoid \cite{malik2007new}, chaotic \cite{feng2007chaotic}, oscillating \cite{kentzoglanakis2009particle}, logarthimic \cite{gao2008particle}, and exponential \cite{li2009particle}. For updating the acceleration coefficients, there are also several techniques inspired mainly by the updating techniques of the inertia weight. 
Based on \cite{bansal2011inertia}, The linear decreasing updating type achieved the minimum error among the rest of the techniques. 
In our proposed model, due to the integration of BPSO within the ANFIS architecture, targeting the minimum error to find the best set of rules is our goal. For this reason, we have applied the technique proposed in \cite{bansal2011inertia} (see equation \ref{eqinertia}), such as 
\cite{liu2015analysis} \cite{wadhawan2022ebpso}. 

    \begin{equation}
    w(t) = w_{\text{max}} - (w_{\text{max}} - w_{\text{min}}) \times \frac{t}{T} 
    \label{eqinertia}
    \end{equation}
The value of $w_{max}$ and $w_{min}$ is 0.9 and 0.4 respectively \cite{he2016parameters}
This updating was conducted by linearly updating the acceleration coefficients  \cite{ardizzon2015adaptive} \cite{tripathi2007multi} based on the swarm performance change in each iteration. 
   
\begin{equation}
    c_{1} = \left( c_{1}\text{max} - c_{1}{\text{min}} \right) \frac{t}{T} + c_{1}{\text{min}}
    \label{eqc1}
    \end{equation}
    \begin{equation}
    c_{2} = \left( c_{2}{\text{max}} - c_{2}{\text{min}} \right) \frac{t}{T} + c_{2}{\text{min}}
    \label{eqc2}
    \end{equation}

where \( t \) denotes the current iteration and \( T\) the maximum number of admissible iterations. The maximum and minimum values for $c_1$ are 2.5 and 0.5, respectively. $c_2$ has the opposite: its maximum value is 0.5, and its minimum value is 2.5  \cite{ardizzon2015adaptive}.

\begin{algorithm*}[h]
\caption{ANFIS-PCA-BPSO}
\footnotesize
\KwData{Input datasets}
\KwResult{Achieving the target, Classification or Regression}

\textbf{Initialization:}
\begin{itemize}
  \item Generate the initial FIS till layer 3 (Normalized firing strengths ($N_{fs}$))
  \item Calculate the number of 
  components based on the explained variance threshold
  \item Perform PCA on firing 
  strengths to reduce dimensionality to $D_{\text{reduced}}$ equal to the number of components
  \item Pass $D_{\text{reduced}}$ to BPSO-based Feature Selection Algorithm 
\end{itemize}

\textbf{Optimization Loop:}
\While{not converged}{
  \For{each particle $i = 1$ to $N$}{
    \For{each dimension $j = 1$ to $D_{\text{reduced}}$}{
      \begin{itemize}
        \item Update the velocity of particle $i$ in dimension $j$ using BPSO equations
        \item Convert the updated velocity to Probability
        \item Update the position of particle $i$ in dimension $j$ based on the probability
        \item Evaluate the fitness of the new position of particle $i$
        \item Update personal best position $P_{best}$ and global best position $G_{best}$ if necessary
      \end{itemize}
    }
  }
}

\textbf{Finalization:}
\begin{itemize}
  \item Report the final FIS with the minimum error for testing
  \item Evaluate the best FIS on the test dataset
  \item Repeat for N folds and take the average performance
\end{itemize}

\end{algorithm*}

\section{Experimental Setup}
In this section, we present the setup of our experiments, such as the dataset used and the evaluation metrics.
\subsection{Dataset}
To assess the performance of our model, we conducted training and evaluation on a set of $8$ classification benchmarks and $4$ regression benchmarks sourced from UCI machine learning respiratory datasets (available at \url{https://archive.ics.uci.edu/ml/index.php}) and keel dataset (available at  \url{https://sci2s.ugr.es/keel/datasets.php}). These datasets encompass a diverse range of classification and regression tasks, with the number of features varying from 2 to 8. Table \ref{table1} provides a detailed description of the selected datasets, including information on the number of features, samples, and classes. 

\begin{table*}[H]
\caption{Detailed Characteristics and Description of the 12 Standard Dataset Used in the Model}
\label{table1}
\footnotesize
\centering
\setstretch{1}
\setlength\tabcolsep{4pt} 
\begin{tabularx}{\textwidth}{p{3cm} p{1.8cm} p{1.8cm} p{1.8cm} p{1.8cm} p{2cm}} 
    \hline
    \textbf{Data title} & \textbf{Abbreviation} & \textbf{\#Features} & \textbf{\#Classes} & \textbf{\#Instances} & \textbf{Task} \\
    \hline
    Iris & IRS & 4 & 3 & 150 & Classification \\
    Teaching Assistant Evaluation & TAE & 5 & 3 & 151 & Classification \\
    Phenome & PHO & 5 & 2 & 5,404 & Classification \\
    Banana & BAN & 2 & 2 & 5300 & Classification \\
    Haberman & HAB & 3 & 2 & 306 & Classification \\
    NewThyroid & THY & 5 & 3 & 215 & Classification \\
    Balance & BAL & 4 & 3 & 625 & Classification \\
    Monk2 & MOK & 6 & 2 & 432 & Classification \\
    Servo & SER & 4 & - & 167 & Regression \\
    Airfoil Noise & AIR & 4 & - & 1503 & Regression \\
    Istanbul Stock Exchange & IST & 8 & - & 536 & Regression \\
    Tecator & TEC & 4 & - & 6000 & Regression \\
    \hline
\end{tabularx}
\end{table*}

\subsection{Evaluation metrices}
This section elucidates the evaluation metrics adopted for gauging our models' performance. We elaborate on the evaluation metrics employed to gauge the performance of our models, particularly in classification tasks. Primarily, we utilize Accuracy,  capturing the ratio of correctly predicted instances. This metric is inherently categorical, aligning with the discrete nature of classification outcomes. Accuracy quantifies the proportion of correctly predicted instances to the total instances, providing a straightforward and intuitive measure of a model's performance. Expressed as a percentage, it offers a clear snapshot of how often the model makes the correct predictions.\cite{chicco2021coefficient}. The mathematical equation for accuracy is shown in equation \ref{eq5.1}.
\begin{equation}
 Accuracy=\frac{TP+TN}{TP+TN+FP+FN}
 \label{eq5.1}
 \end{equation}
Additionally, we incorporate Precision, Recall, and F1-Score to provide a more nuanced assessment. Precision, defined as the ratio of correctly predicted positive observations to the total predicted positives, is crucial in contexts where the cost of false positives is significant; its mathematical equation is represented in equation \eqref{eq5.2}. On the other hand, Recall, or Sensitivity, measures the ratio of correctly predicted positive observations to all actual positives, vital in scenarios where missing a positive instance is particularly consequential, as can be shown mathematically in equation \eqref{eq5.3}. Lastly, the F1-Score, which harmonizes Precision and Recall, is pivotal in balancing the two, especially in imbalanced datasets. This metric is the harmonic mean of Precision and Recall, offering a single metric that encapsulates both aspects. Its mathematical equation is shown in equation \eqref{eq5.4}\cite{grandini2020metrics}.
  \begin{equation}
\text{Precision} = \frac{TP}{TP + FP}
 \label{eq5.2}
\end{equation}

\begin{equation}
\text{Recall} = \frac{TP}{TP + FN}
\label{eq5.3}
\end{equation}

\begin{equation}
\text{F1-Score} = 2 \times \frac{\text{Precision} \times \text{Recall}}{\text{Precision} + \text{Recall}}
\label{eq5.4}
\end{equation}
where:\\
 $TP$ (True Positives) indicates the correct presence of an attribute in the data \\
 $TN$ (True Negatives) indicates the correct absence of an attribute in a data \\
 $FP$  (False Positives) indicates the wrong presence of an attribute in the data \\
 $FN$ (False Negatives) indicates the wrong absence of an attribute in a data \\
 
While other metrics are available for classification evaluation, we predominantly opt for Accuracy, especially to compare fairly with other rules-based reduction techniques. Many models in the literature report their performance in terms of accuracy, making it a de facto standard for comparison. In this light, our decision to prioritize accuracy ensures that our results remain directly comparable and consistent with prevalent practices in the field.

In contrast, regression tasks predict continuous values, necessitating metrics that measure the deviation of predicted values from the actual ones. Thus, we employ the Mean Square Error (MSE), a popular technique used to evaluate model performance by calculating the average of the squares of the difference between each model output and its desired output; the Root Mean Square Error (RMSE), which authorizes large number deviations and punishes significant errors, providing higher weight than MSE. We also considered the Mean Absolute Error (MAE) as an evaluation metric. MAE calculates the average absolute difference between each model output and its desired output. Finally, the Cosine distance evaluation metric is also included. This calculates the pairwise separation between two observations or vectors, representing, in our case, the predicted and actual output. Additionally, we considered the number of optimized rules generated by our model and the estimated training time as essential factors in the evaluation process \cite{chicco2021coefficient}. The mathematical equations for each of these metrics are provided below equations \ref{equ5.2}- \ref{equ5.6}:

\begin{equation}
MSE = \frac{1}{n} \sum_{i=1}^{n} (\hat{y}_i - y_i)^2
\label{equ5.2}
\end{equation}
 \begin{equation}
RMSE = \sqrt{\frac{1}{n} \sum_{i=1}^{n} (\hat{y}_i - y_i)^2}
\label{equ5.3}
\end{equation}
\begin{equation}
MAE = \frac{1}{n} \sum_{i=1}^{n} |\hat{y}_i - y_i|
\label{equ5.4}
\end{equation}
\begin{equation}
CosDistance = 1 - \text{CosineSimilarity}
\label{equ5.5}
\end{equation}
\begin{equation}
CosSimilarity = \frac{\hat{y}_i \cdot y_i}{\|\hat{y}_i\|_2 \cdot \|y_i\|_2}
\label{equ5.6}
\end{equation}

where $n$ is the number of samples, $y_i$  is the actual value of the target variable for the i\textsuperscript{th}  sample, and $\hat{y}_i$ is the predicted value of the target variable for the i\textsuperscript{th}  sample.
 
 Furthermore, we discuss the number of generated rules and computational time as indicators of model efficiency and scalability. These metrics, rooted in the distinct characteristics of classification and regression, provide a holistic perspective on our models' efficacy.

 \section{Results}
 The datasets used in our experiments were split into 80\% for training, and 20\% for testing using 5-fold cross-validation to mitigate overfitting iterating for 100 epochs, and the averages were computed for all evaluation metrics. The membership function type employed was a Generalized Bell shape.  All experiments were conducted on a device with the following specifications: Intel(R) Core(i7) CPU @ 2.70GHz, 12.0 GB of RAM, running 64-bit Windows operating system connected in a home network.

\subsection{Comparing with baseline model}
The first experiment is represented by applying our proposed model, indicated by Figure \ref{fig8}  by applying the PCA as the first stage on the firing strengths. Then, we optimize with further selection using the BPSO. The datasets were preprocessed by splitting into training and testing, normalizing the feature values, and indicating the input and output variables. All the parameters set for this experiment are represented by the number of iterations, which is set to $100$ iterations; the swarm size is equal to the number of generated rules, and the BPSO parameters, $c_1$,$c_2$,$w$, are automatically generated based on swarm performance as explained earlier. 

The model performance is compared concerning the baseline model, the standard (conventional ANFIS) in terms of the accuracy and the number of generated rules, and the training time for the dataset related to the classification purposes and in terms of MSE, MAE, RMSE, and CosDistance for the dataset related to the regression purposes. Table \ref{table3} shows the results achieved by our proposed model in terms
of Accuracy, number of rules, and training time, while Table \ref{table_pre} shows the results of
the precision, recall, and F1-score for the classification task, using the standard dataset. Table \ref{table4} represents the model's performance for the regression dataset compared with the baseline model. 
\begin{table*}[H]
\caption{Classification Performance comparison of our proposed model Across different datasets concerning the baseline model, where Ts\_Acc is the testing accuracy, $\#rules$ is the number of generated rules, and Tr\_time is the training time.}
\label{table3}
\footnotesize
\centering
\setstretch{1}
\begin{tabular}{l|p{4cm}|p{4cm}|p{4cm}} 
\toprule
\textbf{Data} & \textbf{Evaluation Metrics} & \multicolumn{2}{c}{\textbf{Model}} \\
\cmidrule{3-4}
& & \textbf{ANFIS}  & \textbf{ANFIS-PCA-BPSO} \\
\midrule
IRS & Ts\_Acc ($\pm $Std) & 0.953($\pm$0.018)  & \textbf{0.960 ($\pm$0.027)} \\
& \#rules ($\pm$ Std)& 81 &  \textbf{2($\pm$1.22)} \\
& Tr\_time(sec)($\pm$ Std) & 3.53e+03($\pm$6.65e+03)
 &\textbf{13.12($\pm$0.75)}  \\

\midrule
TAE & Ts\_Acc ($\pm $Std) & \textbf{0.583($\pm$0.097)} & 0.542($\pm $0.072)\\
& \#rules ($\pm$ Std)& 32  & \textbf{1.8($\pm $1.3)} \\
& Tr\_time(sec)($\pm$ Std) & 90.7496 ($\pm$10.7209) & \textbf{2.54($\pm$0.34)} \\

\midrule
PHO& Ts\_Acc ($\pm $Std) & \textbf{0.847($\pm $0.0042)}  & 0.845($\pm $0.0045) \\
& \#rules ($\pm$ Std)& 32 & \textbf{4($\pm $1)}\\
& Tr\_time(sec)($\pm$ Std) & 2.08e+03($\pm $4.15e+03) & \textbf{7.6($\pm $0.547)} \\

\midrule
BAN & Ts\_Acc($\pm $Std)  &0.883($\pm $0.0221)  & \textbf{0.891($\pm $0.008)}\\
& \#rules($\pm $Std) & 9  & \textbf{2.2($\pm $1.303)} \\
& Tr\_time(sec)($\pm $Std)  & 336.40($\pm $27.13) & \textbf{7.835($\pm $1.213)} \\

\midrule
HAB & Ts\_Acc($\pm $Std)  & 0.703($\pm $0.0995)  & \textbf{0.745($\pm $0.0744)} \\
& \#rules($\pm $Std) & 27  &\textbf{3.4($\pm $1.3416)}  \\
& Tr\_time(sec)($\pm $Std)  & 114.005($\pm $5.12)  & \textbf{4.9($\pm $0.6922)} \\

\midrule
THY & Ts\_Acc($\pm $Std)  &\textbf{0.930($\pm $0.023)}   & 0.923($\pm $0.0208) \\
& \#rules($\pm $Std) & 32  & \textbf{2.6($\pm $1.34)} \\
& Tr\_time(sec)($\pm $Std)  & 133.095($\pm $25.98) & \textbf{1.8($\pm $0.447)} \\

\midrule
BAL & Ts\_Acc ($\pm $Std) & \textbf{0.886($\pm $0.0203)}  & 0.860($\pm $0.0126) \\
& \#rules($\pm $Std) & 16  &\textbf{3.2($\pm $1.3) }\\
& Tr\_time(sec)($\pm $Std)  & 94.501($\pm $10.006) & \textbf{4.32($\pm $1.58)} \\

\midrule
MOK & Ts\_Acc($\pm $Std)  & 0.991($\pm $0.0097)  & \textbf{1($\pm $0.0)} \\
& \#rules($\pm $Std) & 64  & \textbf{4($\pm $0.0)} \\
& Tr\_time(sec)($\pm $Std)  & 3.17e+03 (2.882e+3) & \textbf{49.62($\pm $7.76)} \\
\bottomrule
\end{tabular}
\end{table*}

\begin{table}[H]
\centering
\setstretch{1}
\caption{Classification Performance comparison of various models on different datasets concerning the baseline model, in terms of other evaluation metrics.}
\label{table_pre}
\footnotesize
\begin{tabular}{p{2cm}|p{3cm}| p{3cm}| p{3cm}}
\toprule
\textbf{Data} & \textbf{Evaluation Metrics} & \textbf{ANFIS} & \textbf{ANFIS-PCA-BPSO} \\
\midrule
\multirow{3}{*}{IRS} & Precision & 0.9622($\pm$0.0418)  & 0.9549($\pm$0.0494) \\
                     & Recall    & 0.9600($\pm$0.043)   & 0.9544($\pm$0.0568) \\
                     & F1-score  & 0.9595($\pm$0.043)   & 0.9523($\pm$0.0545) \\
\hline
\multirow{3}{*}{TAE} & Precision & 0.5822($\pm$0.1203)   & 0.5739($\pm$0.0977) \\
                     & Recall    & 0.5696($\pm$0.1155)   & 0.5571($\pm$0.086) \\
                     & F1-score  & 0.5624($\pm$0.118)   & 0.5534($\pm$0.084) \\
\hline
\multirow{3}{*}{PHO} & Precision & 0.8826($\pm$0.0139)  & 0.8861($\pm$0.0136) \\
                     & Recall    & 0.9005($\pm$0.0090)  & 0.8861($\pm$0.0125) \\
                     & F1-score  & 0.8914($\pm$0.0055)  & 0.8860($\pm$0.0084) \\
\hline
\multirow{3}{*}{BAN} & Precision & 0.9242($\pm$0.0532)   & 0.8645($\pm$0.0893) \\
                     & Recall    & 0.8267($\pm$0.1508)  & 0.8931($\pm$0.0806) \\
                     & F1-score  & 0.8624($\pm$0.0700)  & 0.8723($\pm$0.0285) \\
\hline
\multirow{3}{*}{HAB} & Precision & 0.7515($\pm$0.0237)   & 0.7343($\pm$0.0410) \\
                     & Recall    & 0.9022($\pm$0.0334)        & 0.9822($\pm$0.0100) \\
                     & F1-score  & 0.8197($\pm$0.0225)  & 0.8399($\pm$0.0270) \\
\hline
\multirow{3}{*}{THY} & Precision & 0.9372($\pm$0.0311)   & 0.9205($\pm$0.059) \\
                     & Recall    & 0.8495($\pm$0.069)    & 0.8415($\pm$0.0660) \\
                     & F1-score  & 0.8748($\pm$0.052)    & 0.8690($\pm$0.0625) \\
\hline
\multirow{3}{*}{BAL} & Precision & 0.6038($\pm$0.008)   & 0.6355($\pm$0.0103) \\
                     & Recall    & 0.6389($\pm$0.019)   & 0.6815($\pm$0.0166) \\
                     & F1-score  & 0.6209($\pm$0.0135)  & 0.6557($\pm$0.0129) \\
\hline
\multirow{3}{*}{Mok} & Precision & 0.9903($\pm$0.0134)      & 1($\pm$0.0)         \\
                     & Recall    & 0.9655($\pm$0.0328)         & 1($\pm$0.0)         \\
                     & F1-score  & 0.9774($\pm$0.0170)        & 1($\pm$0.0)         \\
\bottomrule
\end{tabular}
\end{table}

\begin{table*}[H]
\centering
\caption{Regression Performance Comparison of the Proposed Model Across Different Datasets Against the Baseline Model (Cos refers to Cosine Distance)}
\label{table4}
\footnotesize
\setstretch{1}
\begin{tabular}{l|p{3.5cm}|p{4cm}|p{4cm}} 
\toprule
\textbf{Data} & \textbf{Evaluation Metrics} & \multicolumn{2}{c}{\textbf{Model}} \\
\cmidrule{3-4}
& & \textbf{ANFIS}  & \textbf{ANFIS-PCA-BPSO} \\
\midrule
SER & MSE($\pm$Std) & \textbf{0.0126($\pm$0.004)}  & 0.0468($\pm$0.075) \\
& MAE($\pm$Std)& 0.0821($\pm$0.008) &  \textbf{0.0723($\pm$0.041)} \\
& RMSE($\pm$Std) &\textbf{0.1112($\pm$0.0185)}   & 0.1703($\pm$0.149) \\
& Cos($\pm$Std)& \textbf{0.0746($\pm$0.0255)}  & 0.1440($\pm$0.175) \\
& \#rules($\pm$Std) & 16  & \textbf{2.4($\pm$1.14)} \\
& Time($\pm$Std) & 25.530($\pm$2.398)  & \textbf{0.886($\pm$0.121)} \\
\midrule
AIR &  MSE($\pm$Std) &0.0122($\pm$3.82e-04)&	\textbf{0.0092($\pm$0.0013) }\\ 
&MAE($\pm$Std)&0.0851($\pm$0.001)&		\textbf{0.0720($\pm$0.00471)}\\
& RMSE($\pm$Std)	&0.1108($\pm$0.0017)&		\textbf{0.0950($\pm$0.006)}\\
&Cos($\pm$Std)&	0.0170($\pm$5.21e-04)&\textbf{0.0129($\pm$0.0022)}\\
&\#rules($\pm$Std)&	32&	\textbf{2.2($\pm$0.4472)}\\
&Time($\pm$Std)&	2.81e+03($\pm$4.27e+03)&\textbf{23.44($\pm$3.578)}\\
\midrule
IST & MSE($\pm$Std)&	\textbf{0.0033($\pm$8.047e-04)}	&	\textbf{0.0033($\pm$7.666e-04)}\\
&MAE($\pm$Std)&	0.0431($\pm$0.0061)&	\textbf{0.0429($\pm$0.005)}\\
&RMSE($\pm$Std)&	\textbf{0.0567($\pm$0.0071)}&	0.0568($\pm$0.006)\\
&Cos($\pm$Std)&	\textbf{0.0072($\pm$0.0017)}&	\textbf{0.0072($\pm$0.001)}\\
&\#rules($\pm$Std)&	256&	\textbf{2.2($\pm$1.09)}\\
&Time($\pm$Std)&	4.901e+03(4.297e+03)&\textbf{277.161 ($\pm$12.8)}\\
\midrule
TEC & MSE($\pm$Std)&	\textbf{3.90e-04($\pm$3.07e-05)}&4.73e-04($\pm$1.37e-04)\\
&MAE($\pm$Std)&	0.0087($\pm$2.47e-04)&		\textbf{0.0085($\pm$8.13e-04)}\\
&RMSE($\pm$Std)&	\textbf{0.0197($\pm$7.70e-04)}&		0.0216($\pm$0.0031)\\
&Cos($\pm$Std)&\textbf{0.0018($\pm$1.706e-04)}	&		0.0022($\pm$6.11e-04)\\
&\#rules($\pm$Std)&	16&		\textbf{3.2($\pm$1.64)}\\
&Time($\pm$Std)&	2.83e+03($\pm$4.43e+03)&		\textbf{15.03($\pm$3.8)}
 \\
\bottomrule
\end{tabular}
\end{table*}

\subsection{Comparing with state-of-the-art}
To check the validity of our model, we compared it with several state-of-the-art rules-based reduction techniques. For classification, we compared with $4$ techniques aimed to reduce the number of generated rules; they are CFBLS \cite{feng2020accuracy}, D-MOFARC \cite{fazzolari2014multi}, FARC-HD \cite{alcala2011fuzzy}, and PAES-RGT \cite{antonelli2016multi}, all these techniques described Section \ref{related2}. We selected only the common dataset (focusing on the low dimensional dataset with up to 8 features) to present in this study, and these studies evaluated their models mainly based on accuracy and number of generated rules, which we will show in Tables \ref{table6} and \ref{table7}, respectively. The symbol / means this reference did not use this dataset.

\begin{table*}[H]
    \centering
    \caption{Classification Accuracy Comparison of the Proposed Model Across different Datasets Against State-of-the-Art Techniques}
    \label{table6}
    \footnotesize
    \setstretch{1}
    \begin{tabular}{l|p{2cm}|p{2cm}|p{2cm}|p{2cm}|p{2.5cm}}
        \hline
       \multirow{2}{*}\textbf{Data} & \multicolumn{5}{c}{\textbf{Model}} \\
        \cline{2-6}
 & CFBLS \cite{feng2020accuracy} & D-MOFARC\cite{fazzolari2014multi} & FARC-HD \cite{alcala2011fuzzy} & PAES-RGT\cite{antonelli2016multi}  &ANFIS-PCA-BPSO
 \\ \hline 
        IRS & \textbf{0.9822 ($\pm$0.003)} & 0.96 ($\pm$0.0419) & 0.953 & 0.9507 & 0.96 ($\pm$0.027) \\
        TAE & \textbf{0.6406 ($\pm$0.0092)} & 0.59($\pm$0.1179) &0.59 & 0.5618 & 0.57($\pm$0.0609) \\
        PHO & / & 0.835 & 0.824 & 0.8061 & \textbf{0.845($\pm $0.0045)} \\
        BAN & / & \textbf{0.89} & 0.855 & 0.6277 & \textbf{0.891 ($\pm$0.008)} \\
        HAB & 0.7354 ($\pm$0.066) & 0.6940 ($\pm$0.0506) & 0.735 & 0.7426 &\textbf{0.7448 ($\pm$0.0744)} \\
        THY & 0.9508 ($\pm$0.007) & \textbf{0.9550 ($\pm$0.0455) }& 0.941 & 0.9426 & 0.917 ($\pm$0.0596) \\
        BAL & 0.9066 ($\pm$0.003) & 0.8560 ($\pm$0.0326) & \textbf{0.912} & / &0.86($\pm$0.0126) \\
        MOK & 0.9066 ($\pm$0.014) & / & / & \textbf{1}  &\textbf{1($\pm$0.0)} \\
        \bottomrule
    \end{tabular}
\end{table*}

\begin{table*}[H]
    \centering
    \setstretch{1}
    \caption{Comparison of the Number of Rules in the Proposed Model Across Various Classification Benchmarks Against State-of-the-Art Techniques}
    \label{table7}
    \footnotesize
    \setstretch{1}
    \begin{tabular}{l|p{2cm}|p{2cm}|p{2cm}|p{2cm}|p{2.5cm}}
        \hline
       \multirow{2}{*}\textbf{Data} & \multicolumn{5}{c}{\textbf{Model}} \\
        \cline{2-6}
 & CFBLS \cite{feng2020accuracy} & D-MOFARC \cite{fazzolari2014multi} & FARC-HD \cite{alcala2011fuzzy} & PAES-RGT \cite{antonelli2016multi} & ANFIS-PCA-BPSO
 \\ \hline 
        IRS & 5.0 & 5.6 & 4.4 & 19.9  &\textbf{2($\pm$1.22)} \\
        TAE & 16.3 & 20.2 & 19.9 & 22.5 & \textbf{12.4($\pm$3.78)} \\
        PHO & / & 9.3 & 17.2 & 28.4 & \textbf{6.6($\pm$1.67)} \\
        BAN & / & 8.7 & 12.9 & 46.9 &\textbf{2.6($\pm$0.8944) }\\
        HAB & \textbf{2} & 9.2 & 5.7 & 17 & 3.4($\pm$1.3416) \\
        THY & 10 & 9.5 & 4.9 & 18.5 &\textbf{4.4($\pm$2.5)}  \\
        BAL & 9.2 & 19.8 & 18.8 & / &\textbf{3.2($\pm$1.3)} \\
        MOK & 5.9 & / & / & 13.7 &\textbf{4($\pm$0.0)} \\
        \bottomrule
    \end{tabular}
\end{table*}

For regression, we found two main techniques focused on rules-reduction for regression purposes: ANFIS-T \cite{tomasiello2022fractional}, and R-KMELANFIS \cite{pramod2021k} with two attempts (Euclidean and Cosine), also described in Section \ref{related2}. We also selected only the common datasets for our work ( focusing on low-dimensional datasets with up to 8 features). These studies attempt to evaluate and compare their models using the RMSE and the number of generated rules. Table \ref{table10} shows the comparative results of our proposed model compared with these two techniques in terms of RMSE and the number of generated rules.

\begin{table*}[H]
\centering
\footnotesize
\setstretch{1}
\caption{Regression Performance Comparison on Benchmark Datasets with State-of-the-Art Techniques (where \#R denotes the Number of Rules)}
\label{table10}
\begin{tabular}{p{0.7cm}|p{2cm}|p{2cm}|p{2cm}|p{2cm}|p{2.5cm}}
\toprule
Data&Evaluation Metrics & Euclidean-R-KMELANFIS \cite{pramod2021k} & Cosine-R-KMELANFIS \cite{pramod2021k} & ANFIS-T \cite{tomasiello2022fractional} & ANFIS-PCA-BPSO\\
\midrule
\multirow{2}{*}{SER} & RMSE ($\pm$Std) & 0.3237 ($\pm$0.0554) & 0.3992 ($\pm$0.0832) & 0.1775 ($\pm$0.0198)  & \textbf{0.1703 ($\pm$0.149)} \\
& \#R & 6 & 9 & 3 &  \textbf{2.4 ($\pm$1.14)} \\\hline
\multirow{2}{*}{AIR} & RMSE ($\pm$Std) & 3.1799 ($\pm$0.143) & 3.6996 ($\pm$0.141) & 4.9726 ($\pm$0.5354)  &\textbf{0.095 ($\pm$0.006)}  \\
& \#R & 38 & 30 & 4  &\textbf{2.2 ($\pm$0.447)}  \\\hline
\multirow{2}{*}{IST} & RMSE ($\pm$Std) & 0.012 ($\pm$0.00071) & 0.012 ($\pm$0.00071) & \textbf{0.00482 ($\pm$0.00049)}  & 0.0568 ($\pm$0.006) \\
& \#R & \textbf{2} & \textbf{2} & 4  & 2.2 ($\pm$1.09) \\\hline
\multirow{2}{*}{TEC} & RMSE ($\pm$Std) & 0.3854 ($\pm$0.020) & 0.3945 ($\pm$0.014) & 0.3351 ($\pm$0.00487)  & \textbf{0.0216 ($\pm$0.0031)} \\
& \#R & 6 & 9 & \textbf{3} & 3.2 ($\pm$1.64) \\
\bottomrule
\end{tabular}
\end{table*}

\section{Ablations}
In our ablation study, we systematically investigated several key issues to demonstrate the impact of each addition to the original ANFIS architecture, ultimately leading to our final proposed model. This section aims to elucidate the rationale and outcomes behind the various modifications and enhancements implemented during our research.

\textbf{Investigating Feature Reduction Techniques} 
Selecting an appropriate feature reduction technique is crucial and dependent on the type of data and the specific application. Selecting Principal Component Analysis (PCA) for feature reduction in our work, as opposed to other techniques such as Linear Discriminant Analysis (LDA), Generalized Discriminant Analysis (GDA), Singular Value Decomposition (SVD), t-distributed Stochastic Neighbor Embedding (t-SNE), and Non-negative Matrix Factorization (NMF), is based on several key considerations. While SVD is a robust method for data decomposition, PCA is preferred due to its focus on variance, interpretability, and computational efficiency, particularly in scenarios where the primary goal is featuring reduction, and this is one of our goals while searching. PCA’s unsupervised nature makes it applicable to a broader range of datasets, unlike LDA and GDA, which require labeled data. This versatility is crucial for my analysis, which includes datasets without predefined classes. Additionally, PCA stands out for its simplicity and interpretability, offering a more straightforward understanding of the transformed feature space compared to the more complex methodologies of SVD, t-SNE, or NMF. Moreover, PCA’s computational efficiency is particularly important when dealing with large datasets, avoiding the high computational costs associated with methods like t-SNE and SVD. Its linear transformation approach effectively captures variance in data, making it suitable for datasets where linear relationships among features are prevalent. Unlike t-SNE, which focuses on local structures and is mainly used for visualization, PCA maintains the global structure of the data, essential for our analytical tasks. Also, unlike NMF, PCA does not impose non-negativity constraints on the data, making it applicable to a wider variety of datasets, including those with negative values. The proven effectiveness of PCA across various domains further solidifies its suitability for our study, offering a reliable and tested approach for dimensionality reduction.

\textbf{Investigating using only BPSO for rules reduction}
In our previous publication \cite{al2023predicting}, we explored the use of Binary Particle Swarm Optimization (BPSO) solely for feature selection, which indirectly leads to rules reduction. Although this approach did reduce the number of rules, it remained relatively high compared to other state-of-the-art techniques that rely on more precise initialization of selected rules.

\textbf{Investigating using only PCA for rules reduction}
We also experimented with using PCA alone for rules reduction, but this method proved insufficient for several reasons:
\begin{itemize}
    \item Integrating PCA alone with ANFIS resulted in a high number of generated components based on data variability, leading to an excessive number of rules.
\item The output of PCA required extensive training and evaluation within the ANFIS architecture to ensure optimal model performance, which involved time-consuming backpropagation processes. 
\item Despite the integration, the model's performance metrics (accuracy, precision, recall, and F1-score) were lower compared to our proposed model. This suggests an incompatibility between the hybrid training of ANFIS with PCA, indicating the need for modification or replacement with other techniques.
\end{itemize}
By addressing these issues through our ablation studies, we have refined our approach to enhance the efficiency and effectiveness of our proposed ANFIS model.

\section{Discussion}
Compared with the baseline model, our proposed model, ANFIS-PCA-BPSO manifested two salient advancements: a marked reduction in training time and rule generation, pivotal metrics in model efficiency, and computational expenditure. An empirical observation, as per Table \ref{table3}, indicates that reducing the number of generated rules and training time across all datasets is beneficial in mitigating the computational and temporal overheads often associated with machine learning model training.

ANFIS-PCA-BPSO consistently outperformed in reducing the number of generated rules and training time across various datasets. Integrating PCA and BPSO within the ANFIS architecture is pivotal. PCA is renowned for transforming original variables into a new set of uncorrelated variables (principal components), which retain most of the data's variance. Thus, our approach of utilizing PCA for dimensionality reduction in conjunction with BPSO for rule optimization mitigates the issue of excessive rule generation in fuzzy inference systems and ensures that the most significant rules, in terms of data variance, are retained, thereby preserving predictive integrity.

However, it is imperative to address an observed trade-off. While our model significantly reduces rule generation and computational time, a nuanced decrease in accuracy exists in certain datasets, such as TAE, PHO, THY, and BAL. This phenomenon is emblematic of the well-established bias-variance trade-off in machine learning, where a reduction in model complexity (via rule reduction, in this context) can occasionally induce an increase in bias and a slight diminution in model accuracy. Nevertheless, it is critical to note that this minor attenuation in accuracy is often deemed acceptable in light of the substantial computational and temporal savings, which are particularly significant in real-time and large-scale applications. These
outcomes have been double-approved by the precision, recall, and F1-score shown in
Table \ref{table_pre}, where for most datasets, our proposed model tends to have similar performance across
all three metrics, which suggests that it is neither overly biased towards precision or recall. It can be specifically noticed that the ANFIS model generally shows a higher precision across most datasets compared to the proposed model, which might suggest
that it is better at predicting positive instances. However, this does not always translate
to the highest F1 score, which is a more balanced metric considering both precision and recall. Including PCA does not consistently improve or worsen the performance across all datasets. For instance, in the BAN or HAB dataset, the inclusion of PCA slightly decreases
precision but increases recall. This might indicate that PCA is helping the model to generalize better but at the cost of incorrectly classifying some negative instances as positive.

Moreover, despite this, ANFIS outperformed traditional ANFIS for the rest of the datasets after integrating PCA, underscoring a notable distinction. This can be attributed to our approach of transforming all generated rules, both redundant and significant, and subsequently discarding the least essential ones, which ensures the preservation of significant rules within the initial PCA components, thereby often achieving higher accuracy.

Figure \ref{figclas_1} visually represents the accuracy, training time, and number of rules among the classification benchmarks for the proposed model and the baseline.

\begin{figure} [htbp]
\centering
\includegraphics[width=0.5\linewidth]{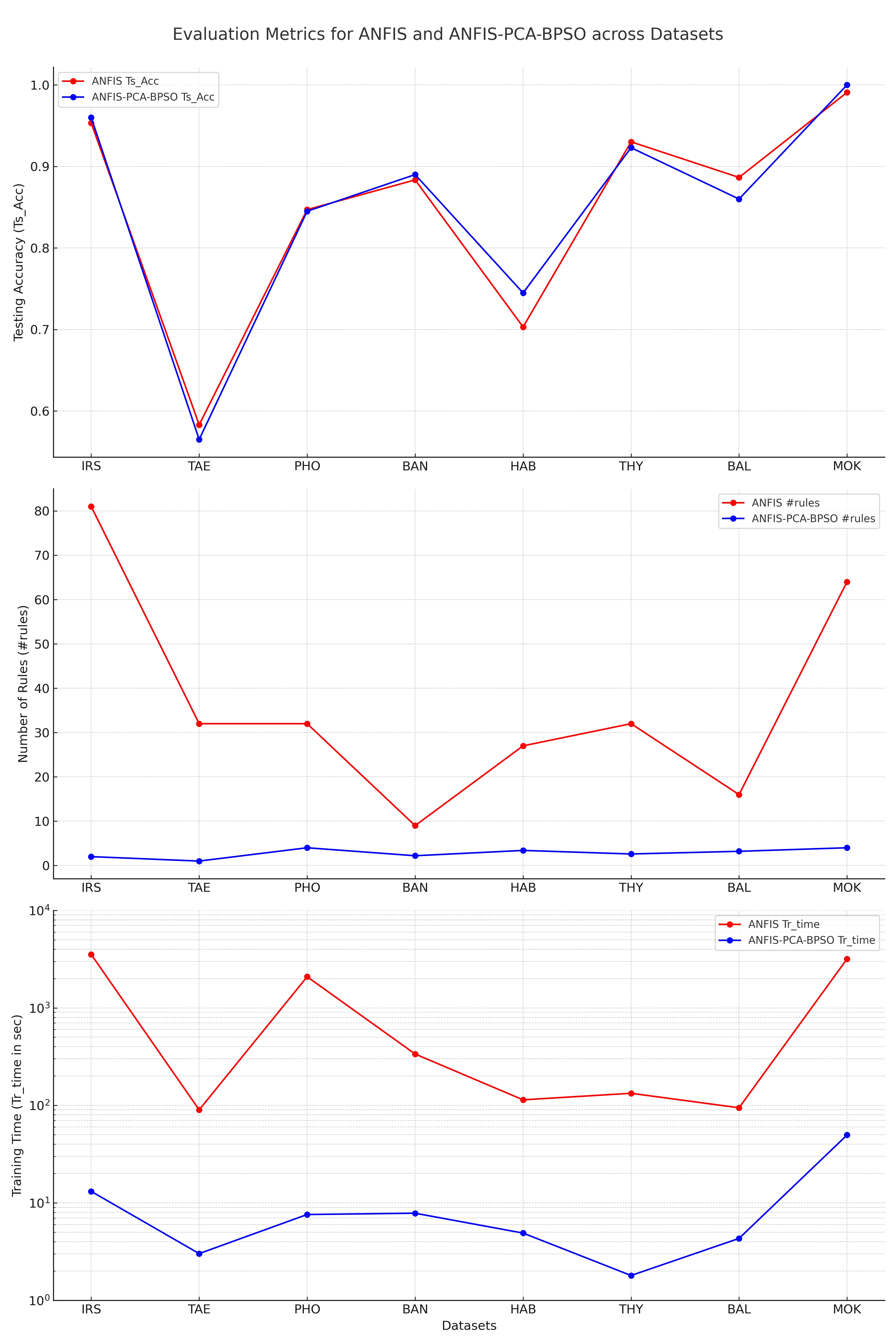}
\caption{Comparative Evaluation of ANFIS and ANFIS-PCA-BPSO Models on Various Classification Datasets.}
\label{figclas_1}
\end{figure}

Regression benchmark datasets further elucidate the significant reduction in training time and rule generation by the ANFIS-PCA-BPSO model. While a slight reduction in some evaluation metrics was observed for certain datasets, such as the MSE, RMSE, and Cos for the SER and TEC datasets, it is acknowledged as an acceptable degradation given the concurrent reduction in computational complexity and training time. Notably, with the AIR dataset, all the evaluation metrics of our proposed model outperformed the traditional ANFIS, and all of them except the RMSE outperformed the traditional ANFIS for the IST dataset.

Figure \ref{figReg_1} delineates the relationship between our proposed model and the traditional ANFIS across these regression benchmarks for all evaluation metrics.

\begin{figure*} [h]
\centering
\includegraphics[width=0.95\linewidth]{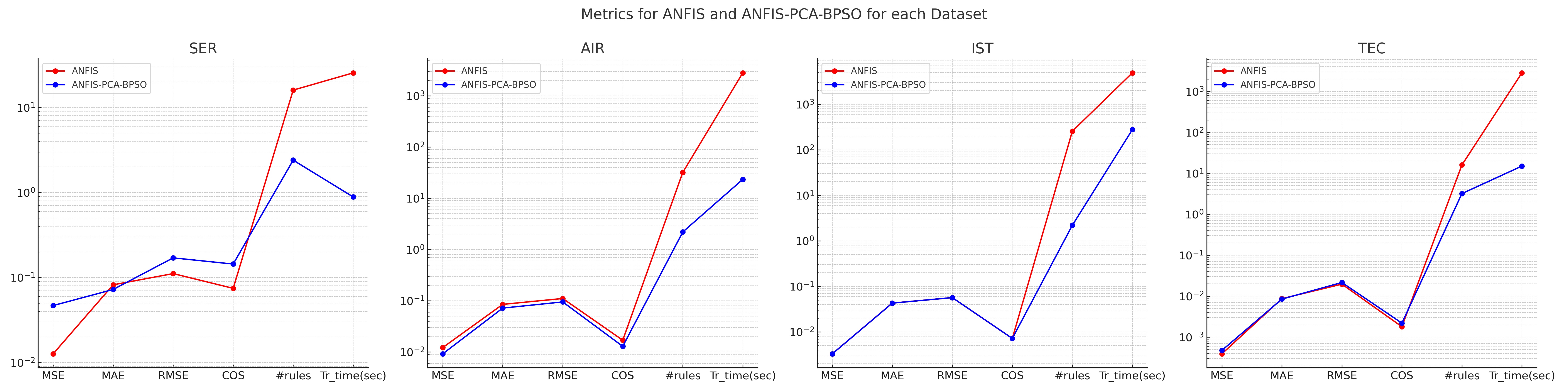}
\caption{Comparative Evaluation of ANFIS and ANFIS-PCA-BPSO Models on Various Regression Datasets.}
\label{figReg_1}
\end{figure*}

Pitted against state-of-the-art rule-based reduction techniques, our models demonstrated competitive, if not superior, performance on several fronts. 

Some methods, specifically CFBLS and DMOFARC, occasionally did better than our model in being accurate. However, when creating rules efficiently, they were still not as good as our model. Regarding training time, an absence of explicit documentation in their respective publications precluded a comparative analysis in this dimension, thereby introducing an element of analytical opacity.

Table \ref{table6} illuminates that ANFIS-PCA-BPSO not only achieved pinnacle accuracy for half of the datasets (4 out of 8) but also minimized rule generation for a commanding majority (7 out of 8) as shown in Table \ref{table7}. Even when traversing datasets where accuracy was not paramount, the performance remained close to alternate models. A deliberate omission of training time from this comparative analysis was necessitated due to the extraction of results for alternative techniques directly from respective publications, wherein training time data remained undisclosed.

Figure \ref{figcls_2} elucidates a comparative canvas, presenting a visual comparison between our propounded model and the state-of-the-art techniques, articulating accuracy and rule generation metrics, and providing a bifocal lens through which model performance can be appraised.

\begin{figure} [h]
\centering
\includegraphics[width=0.7\linewidth]{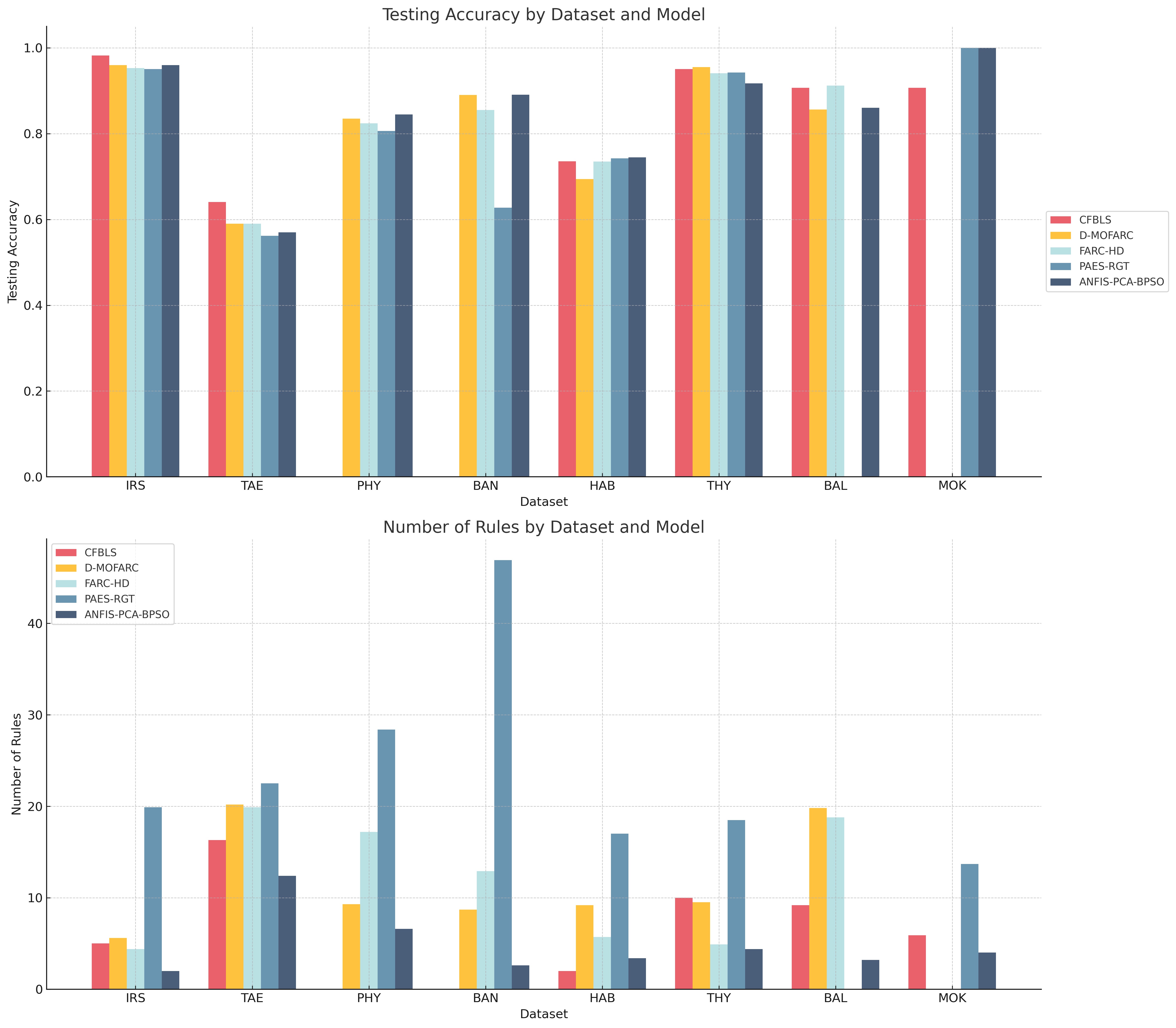}
\caption{A Detailed Analysis of Algorithm Performance: Examining Rule Count and Accuracy Against State-of-the-art Techniques for the Classification Benchmarks.} 
\label{figcls_2}
\end{figure}

Concerning regression-oriented datasets, as illustrated in Table \ref{table10}, our proposed model eclipsed 3 out of 4 of the datasets in terms of the RMSE evaluation metric when compared with alternate techniques. The arena of rule generation presented a more intricate landscape, with ANFIS-PCA-BPSO securing a competitive stance for half of the datasets (2 out of 4). 

Figure \ref{figreg_2} visually elucidates the performance panorama of our proposed model relative to selected state-of-the-art techniques across regression benchmarks, crafting a comprehensive comparative tableau.

\begin{figure} [h]
\centering
\includegraphics[width=0.7\linewidth]{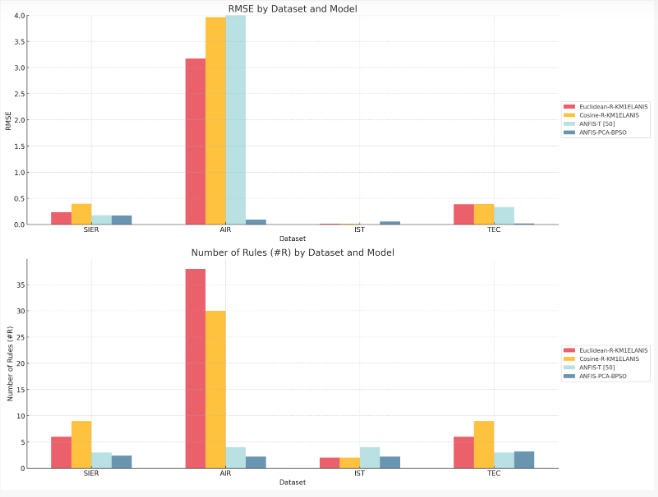}
\caption{A Detailed Analysis of Algorithm Performance: Examining Rule Count and RMSE Against
State-of-the-art Techniques for the Regression Benchmarks.} 
\label{figreg_2}
\end{figure}

\section{Real-World Application}\label{realDes}
The ischemic Stroke dataset has been used in \cite{kamran2017prediction},\cite{ali2018use}, and \cite{al2023predicting}, and it is not publicly available. It consists of $204$ records with $11$ characteristics, approved based on the Neurologist’s opinion. This dataset comprises patient information gathered from a pooled Decompressive Hemicraniectomy database, the components of which were received from three referral centers in three distinct countries, namely Qatar, the United Arab Emirates, and Pakistan. Only patients with three brain computed tomography (CT) scans and signs of acute ischemia were considered. These specifics include the patient’s age, whether they have diabetes, whether they had Hemicraniectomy, their hypertension status, whether they have Dyslipidemia, blood pressure readings, INFARCT VOLUME 1 and 2, and the First infarction growth rate per hour. All these features are described in detail in Table \ref{table2} with their meanings, range of values, the P\_value, and correlation coefficient with our target, the second infarction growth rate (IGRII).
\begin{table*}[H]
\caption{A detail Characteristics and Description of the Real datasets (the Ischemic Stroke Dataset.)}
\label{table2}
\footnotesize
\centering
\setstretch{1}
\setlength\tabcolsep{3.5pt} 
\begin{tabularx}{\textwidth}{p{2cm} p{4cm} p{3cm} p{2cm} p{2.5cm}} 
    \hline
    \textbf{Feature} & \textbf{Description} & \textbf{Values} & \textbf{P\_value} & \textbf{Correlation with IGR2} \\ 
    \hline
    AGE & Age of the patient & in years & 0.8058 & -0.0187 \\
    SBP & Systolic blood pressure & - & 0.9253 & 0.0071 \\
    DBP & Diastolic blood pressure & - & 0.6222 & 0.0374 \\
    HTN & Hypertension diagnosis & 0 – Absent,1 - Present & 0.9350 & 0.0062 \\
    DM & Diabetes Mellitus & 0 – Absent, 1 - Present & 0.3811 & 0.0664 \\
    DYSLIP & Dyslipidemia & 0 – Absent, 1 -Present & 0.1750 & -0.1027 \\
    UNCAL & Uncal Herniation & 0 – Absent, 1 -Present & 0.0312 & 0.1625 \\
    TEMPORAL & Temporal Lobe Involved & 0 – Absent, 1 -Present & 0.0105 & 0.1926 \\
    INFVOL1 & Infarct Volume 1 (CM3) & - & 0.0100 & -0.1936 \\
    INFVOL2 & Infarct Volume 2 (CM3) & - & 6.3804e-09 & 0.4202 \\
    Growthrate\_1 & 1\textsuperscript{st} infarction growth rate/hr & - & 2.1920e-33 & 0.7525 \\
    \hline
\end{tabularx}
\end{table*}

Regarding this dataset, the aim is to predict the Infraction Growth Rate II (IGR II). It is usually calculated after two CT scan rounds, which is time-consuming and cost-effective, and finding an AI model that can predict this important factor after only one CT scan round has the benefit of speeding up the process of diagnosis and saving costs. It represents a regression task, so its evaluation is done based on regression evaluation metrics using a generalized bell-shaped membership function type. We calculated this dataset's P-value and correlation coefficient between each feature and our target (the IGR II). This calculation helped us choose the most significant features that impact the prediction of the IGR II. The final set of features selected for our model is (DYSLIP, UNCAL, TEMPORAL, INVOL1, and Growthrate\_1). For all these features, their p-value is very close to its threshold of 0.05 and the highest correlation coefficient of absolute 0.1. Regarding the 'DYSLIPIDEMIA' feature, based on a study in 2022 \cite{shah2022dyslipidemia}, this feature is a significant risk factor for coronary heart disease, but its impact on ischemic stroke is still under discovery, so having P-value very close to the threshold of P-values motivated us to add this feature to the set of selected features. Those features that need normalization were normalized in the range of 0,1 to unify the range of their values. We excluded INVOL2 (Infarct Volume 2) because this feature can be extracted after the second CT scan round, which is not considered for our study. In addition to the evaluation metrics mentioned earlier for the regression tasks, we added the p-value and the Pearson correlation coefficient between the predicted output and the actual label for medical accuracy purposes for evaluating the real dataset.
\begin{table}[H]
  \centering
  \setstretch{1}
  \caption{Average Regression Evaluation metrics for our proposed model concerning the baseline model on the real dataset using generalized bell membership function.}
  \label{table5}
   \footnotesize
   \begin{tabular}{p{2.3cm}|p{2.7cm}|p{3.5cm}}
    \hline
    \multirow{2}{*}{\textbf{Metrics}} & \multicolumn{2}{c}{\textbf{Model}} \\
    \cline{2-3}
     & \textbf{ANFIS} &\textbf{ANFIS-PCA-BPSO} \\ \hline
    
    MSE ($\pm$ Std) & \textbf{0.0153 ($\pm$ 0.007)} &0.0202 ($\pm$ 0.013) \\
    MAE ($\pm$ Std) & \textbf{0.0813 ($\pm$ 0.016)} &0.0837 ($\pm$ 0.021) \\
    RMSE ($\pm$ Std) & \textbf{0.1207 ($\pm$ 0.031) }&0.1361 ($\pm$ 0.045)\\
    Cos ($\pm$ Std) & \textbf{0.1293 ($\pm$ 0.06)} &0.1607 ($\pm$ 0.071) \\
    p-value & 0.00019 &  \textbf{0.00007} \\
    Correlation & \textbf{0.68279}  &0.6501 \\
    \#rules ($\pm$ Std) & 32  &\textbf{2.4 ($\pm$1.14 ) } \\
    Time ($\pm$ Std) & 256.50 ($\pm$ 20.9) &\textbf{9.4 ($\pm$ 1.09)} \\ \hline
  \end{tabular}
\end{table}

This dataset has been split into 80\% training, and 20\% testing and by using 5-folds cross validation with 150 epochs of training as shown in Table \ref{table5} which presented four evaluation metrics (MSE, MAE, RMSE, CosDistance), statistical analysis (p-value, person correlation), as well as the number of generated rules and the training time. This experiment on the real dataset was conducted using the gbell type of membership function of value being $2$, which is commonly used, and we selected this type as we noticed good results while using it with the classification benchmarks. 

A discerning observation reveals that our proposed model, the ANFIS-PCA-BPSO, surpassed the baseline regarding training time and rule generation, achieving a notable enhancement in computational efficiency and model simplicity. While a subtle discrepancy is observed concerning the remaining evaluation metrics, it is deemed acceptable given the significant reduction in computational complexity and training time. It is paramount in practical, real-world applications where computational resources may be limited or real-time predictions are requisite.

Interestingly, the p-value for ANFIS-PCA-BPSO was significantly lower than the baseline, thereby indicating a robust statistical significance in its predictions and underscoring a high correlation between the predicted IGR II and the baseline, as corroborated by the Pearson correlation coefficient. This implies that ANFIS-PCA-BPSO retains predictive accuracy despite rule and time reductions and provides statistically significant and highly correlated predictions, reaffirming its viability as a predictive model in practical applications. Figure \ref{figReg_base} visualizes the performance of our proposed model and the baseline regarding the mentioned evaluation metrics.
\begin{figure*} [htbp]
  \centering
  \includegraphics[width=0.8\linewidth]{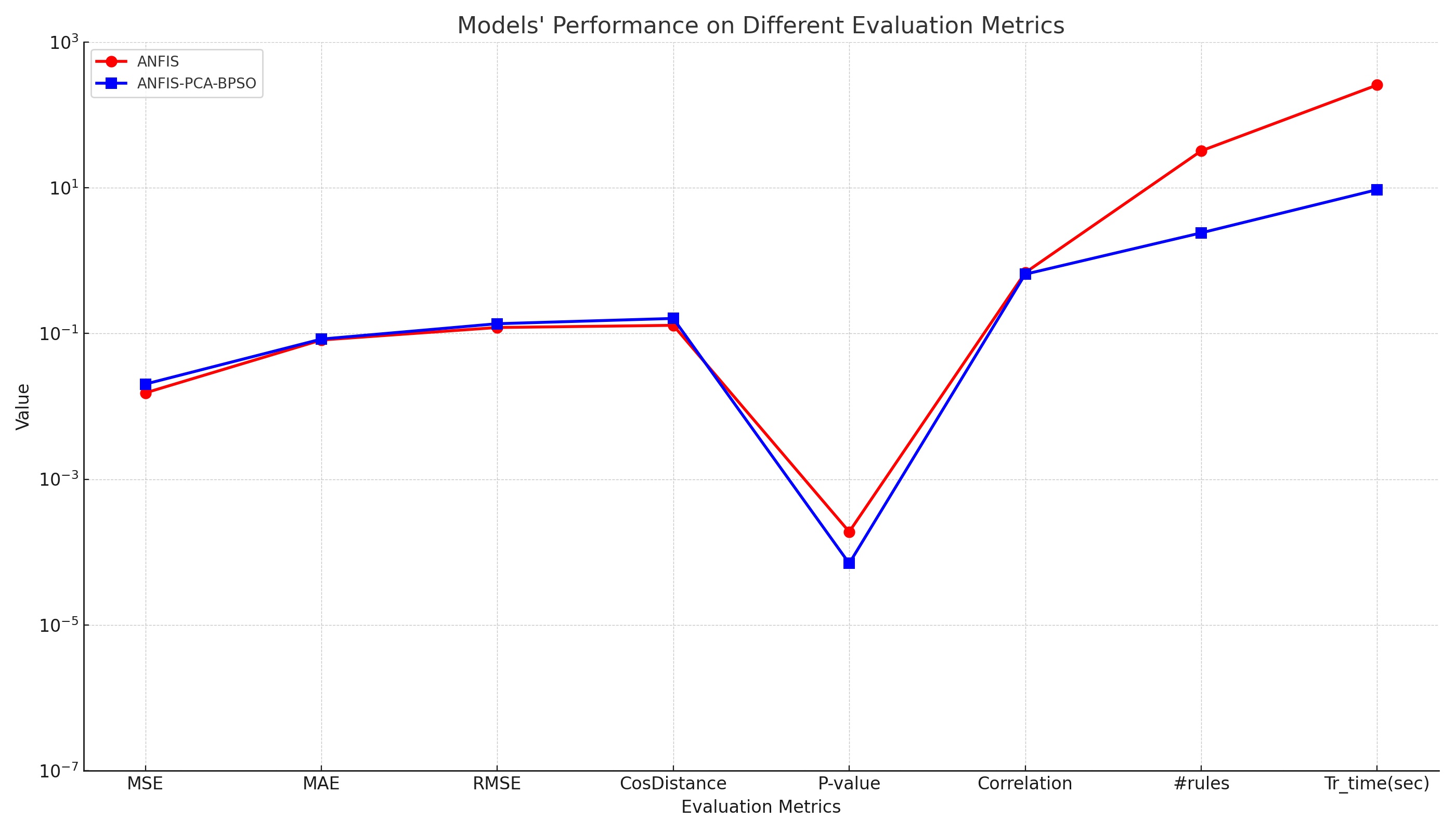}
  \caption{Comparative Evaluation of ANFIS and ANFIS-PCA-BPSO on a Real Dataset Using Multiple Evaluation Metrics.}
  \label{figReg_base} 
\end{figure*} 

As discussed in Section \ref{realDes}, This dataset is not publicly accessible. This restriction prevents us from making a direct comparison with other works. Nevertheless, a comparison was made with two techniques: the approach proposed by \cite{ali2018use}, and this is for two primary reasons. Firstly, their research objective is aligned with ours, focusing on the prediction of IGRII. Secondly, they utilized a dataset similar to ours. It is also compared with our primary model ANFIS-BPSO \cite{al2023predicting} using the Generalized bell shape membership function. As shown in Table \ref{table5}, the performance of our proposed model was not as efficient as the traditional Model despite the closeness in their values. As an attempt to improve the proposed model performance, increasing the number of membership function can be useful as more rules will be generated, however, without any guarantee about their significance.  For this reason ,during the comparison with the other techniques , we repeated our experiment when the membership function is 3 , and Table \ref{table8} provides the comparative results considering Root Mean Square Error and Cosine distance, as these were the only evaluation metrics shared between our study and \cite{ali2018use}\cite{al2023predicting}.
\begin{table*}[H]
    \centering
    \setstretch{1}
    \footnotesize
    \caption{Detailed Comparison of Results Between Our Proposed Model, the Baseline Model, and Other Studies used the same real dataset.}
    \label{table8}
    \begin{tabular}{p{2cm} p{5cm} p{2cm} p{2cm}}
        \toprule
        Model & Description & RMSE & CosDistance \\
        \midrule
        \cite{ali2018use} & without PCA & 0.439 & 0.616 \\
        \cite{ali2018use} & with PCA & 0.196 & 0.464 \\
        ANFIS(mf=3)& using highly correlated features & 0.1266  & 0.1293 \\
        ANFIS-BPSO(mf=3) \cite{al2023predicting} & using highly correlated features & 0.1439 & 0.1328 \\
        ANFIS-PCA-BPSO(mf=3) & using highly correlated features & \textbf{0.1186} & \textbf{0.1261}\\
        \bottomrule
    \end{tabular}
\end{table*}
It can be seen that our proposed model and the baseline showcased commendable performance in RMSE and Cosine distance when operating on features exhibiting high correlation with the target IGR II. A nuanced distinction emerges, with conventional ANFIS registering the lowest RMSE of 0.1186 for ANFIS-PCA-BPSO, compared to 0.1439 for ANFIS-BPSO \cite{al2023predicting}. This subtle differentiation illuminates that despite the larger number of generated rules when the membership function has increased, the ANFIS-BPSO \cite{al2023predicting} may exhibit marginal performance fluctuations due to the stochastic nature of rule selection, as contrasted with our PCA-integrated model, which transmutes all rules into a combination, thereby aggregating the most potent rules within the initial N components based on data variability, and the irrelevant components of insignificant rules have been removed. This rule reduction, however, is articulated within a framework of substantially diminished training time, approximately bisected, and a constricted rule generation volume, particularly when contrasted with traditional ANFIS, which indicates that not always increasing the number of rules will lead to high performance and may these generated rules include a large amount of redundant rules which definitely affect on the overall model performance.

The exploration in \cite{ali2018use}, which leveraged PCA for feature reduction, may not invariably generate a comprehensive set of pivotal rules despite its proximate performance to our models. Furthermore, their strategy may incorporate redundant features, potentially attenuating overall model performance. By electing the most salient features and identifying up to five significant features predicated on the p-value and correlation coefficient, we ensured the generation of essential rules for target prediction. Integration with PCA and BPSO further assured rule number reduction while enhancing performance.

\section{Conclusion}
In the presented study, the proposed model ANFIS-PCA-BPSO is explored as a solution to the challenge of excessive rule generation in fuzzy inference systems, specifically within classification and regression tasks. This model not only exhibited competitive accuracy but also efficiently addressed rule proliferation, standing out even when compared to other state-of-the-art rule reduction techniques. While certain models occasionally surpassed ANFIS-PCA-BPSO in terms of accuracy, the significant reduction in rule generation underscores its potential to alleviate computational complexities. The research incorporated strategies like binary particle swarm optimization (BPSO) and principal component analysis (PCA) into the ANFIS framework to prune and optimize the rule set strategically. Despite its strengths, the model's efficacy diminishes with datasets having a large number of features. Overall, these advancements promise a balance between transparency, adaptability, and computational efficiency in rule-based systems, with potential future applications spanning various domains.

\bibliographystyle{elsarticle-num} 
 \bibliography{main}


\label{}


\end{document}